\documentclass[journal abbreviation, manuscript]{copernicus}

\usepackage{rotating}
\usepackage{amsmath}
\usepackage{setspace}

\newcommand{\ulmo}{{\sc Ulmo}}
\newcommand{\enki}{{\sc Enki}}
\newcommand{\ssta}{SSTa}
\newcommand{\LL}{{\ensuremath{\rm LL}}}
\newcommand{\CC}{{\ensuremath{\rm CC}}} % Cloud coverage
\newcommand{\vitmae}{ViTMAE}
\newcommand{\llc}{LLC4320}

\newcommand{\ml}{ML} % Machine Learning
\newcommand{\ai}{AI} % Machine Learning

\newcommand{\tper}{\ensuremath{t}}  % Training percentage
\newcommand{\pper}{\ensuremath{p}}  % Patch percentage
\newcommand{\stdT}{\ensuremath{\sigma_T}}  % Patch percentage

\begin{document}

%% ------------------------------------------------------------------------ %%
%  Title
%
% (A title should be specific, informative, and brief. Use
% abbreviations only if they are defined in the abstract. Titles that
% start with general keywords then specific terms are optimized in
% searches)
%
%% ------------------------------------------------------------------------ %%

\nolinenumbers

\title{Reconstructing Sea Surface Temperature Images: A Masked Autoencoder Approach for Cloud Masking and Reconstruction}

%% ------------------------------------------------------------------------ %%
%
%  AUTHORS AND AFFILIATIONS
%
%% ------------------------------------------------------------------------ %%

% Authors are individuals who have significantly contributed to the
% research and preparation of the article. Group authors are allowed, if
% each author in the group is separately identified in an appendix.)

% List authors by first name or initial followed by last name and
% separated by commas. Use \affil{} to number affiliations, and
% \thanks{} for author notes.
% Additional author notes should be indicated with \thanks{} (for
% example, for current addresses).

% Example: \authors{A. B. Author\affil{1}\thanks{Current address, Antartica}, B. C. Author\affil{2,3}, and D. E.
% Author\affil{3,4}\thanks{Also funded by Monsanto.}}

%\authors{A. Agabin\affil{1}, J. Xavier Prochaska\affil{1}}

\Author[1, $\star$]{Angelina}{Agabin}
\Author[2,3 $\star$]{J. Xavier}{Prochaska}

\affil[1]{Department of Applied Math, University of California, Santa Cruz, CA, 95064, USA}
\affil[2]{Affiliate of the Department of Ocean Sciences, University of California, Santa Cruz, CA, 95064, USA}
\affil[3]{Department of Astronomy and Astrophysics, University of California, Santa Cruz, CA, 95064, USA}

% \affiliation{1}{First Affiliation}
% \affiliation{2}{Second Affiliation}
% \affiliation{3}{Third Affiliation}
% \affiliation{4}{Fourth Affiliation}

%\affiliation{1}{University of California Santa Cruz}

%(repeat as many times as is necessary)

%% Corresponding Author:
% Corresponding author mailing address and e-mail address:

% (include name and email addresses of the corresponding author.  More
% than one corresponding author is allowed in this LaTeX file and for
% publication; but only one corresponding author is allowed in our
% editorial system.)

% Example: \correspondingauthor{First and Last Name}{email@address.edu}

%\correspondence{J. Xavier Prochaska (jxp@ucsc.edu)}

\runningtitle{TEXT}

\runningauthor{TEXT}

\received{}
\pubdiscuss{} %% only important for two-stage journals
\revised{}
\accepted{}
\published{}

%% These dates will be inserted by Copernicus Publications during the typesetting process.

\firstpage{1}

\maketitle

%\vfill
%Approved: \hrulefill

%\hspace*{0mm}\phantom{Approved: }J.X. Prochaska, Ph.D.

%% Keypoints, final entry on title page.

%  List up to three key points (at least one is required)
%  Key Points summarize the main points and conclusions of the article
%  Each must be 100 characters or less with no special characters or punctuation and must be complete sentences

% Example:
% \begin{keypoints}
% \item	List up to three key points (at least one is required)
% \item	Key Points summarize the main points and conclusions of the article
% \item	Each must be 100 characters or less with no special characters or punctuation and must be complete sentences
% \end{keypoints}

%% ------------------------------------------------------------------------ %%
%
%  ABSTRACT and PLAIN LANGUAGE SUMMARY
%
% A good Abstract will begin with a short description of the problem
% being addressed, briefly describe the new data or analyses, then
% briefly states the main conclusion(s) and how they are supported and
% uncertainties.

% The Plain Language Summary should be written for a broad audience,
% including journalists and the science-interested public, that will not have 
% a background in your field.
%
% A Plain Language Summary is required in GRL, JGR: Planets, JGR: Biogeosciences,
% JGR: Oceans, G-Cubed, Reviews of Geophysics, and JAMES.
% see http://sharingscience.agu.org/creating-plain-language-summary/)
%
%% ------------------------------------------------------------------------ %%

%% \begin{abstract} starts the second page

%\linenumbers

\begin{abstract}

This thesis presents a new algorithm to mitigate cloud masking in the analysis of sea surface temperature (SST) data generated by remote sensing technologies, e.g., 
the Level-2 Visible-Infrared Imager-Radiometer Suite (VIIRS). %The 2nd full-mission reanalysis (RAN2) of the VIIRS dataset provides a substantial source of big data (nearly ~100Tb) however 
Clouds interfere with the analysis of
all remote sensing data using wavelengths
shorter than $\approx 12$\,microns, 
significantly limiting the quantity of usable data
and creating a biased geographical 
distribution (towards equatorial and coastal regions).
%, or containing improperly unmasked clouds. 
%Prior studies have led to use of in-painting algorithms like Navier-Stokes but was typically only used up to 5\%\ masking and had limited
%success. 
To address this issue, we propose an unsupervised machine learning algorithm called \enki\ which uses a Vision Transformer with Masked Autoencoding to reconstruct 
masked pixels. 
We train four different models of \enki\ with varying
mask ratios (referred to as \tper) of 10\%, 35\%, 50\%, and 75\%\ on the generated Ocean General Circulation Model (OGCM) dataset referred to as \llc. 
To evaluate performance, we reconstruct a validation
set of \llc\ SST images with random ``clouds''
corrupting \pper=10\%, 20\%, 30\%, 40\%, 50\%\ of 
the images with individual patches of $4 \times 4 \, 
\rm pixel^2$. 
%and examine reconstruction qualitatively and statistically by calculating the root means squared error (RMSE) of reconstructed patches. 
We consistently find that at all levels of \pper\ 
there is one or multiple models that 
reconstruct the images with a mean RMSE of less than 
$\approx 0.03$K, i.e.\ lower than the estimated sensor 
error of VIIRS data. 
%$\approx 0.078$ K for daytime, along scan, and $\approx 0.05$ K for nighttime, along-scan. 
Similarly, at the individual patch level, the 
reconstructions have RMSE $\approx 8\times$ 
smaller than the fluctuations in the patch.
And, as anticipated, reconstruction errors are larger for
images with a higher degree of complexity.
%the complexity of dynamics within an image and the \pper\ masking ratio affect RMSE with higher complexity and \pper\ masking seeing higher RMSE values. 
Our analysis also reveals that patches along the
image border have systematically higher reconstruction
error; we recommend ignoring these in production. 
%and that a bias appears in some models when reconstructing images at \pper\ masking ratios away from their training mask ratio \tper. 
%Critically, we also discover at a patch level that despite RMSE having some correlation to complexity, they are not directly proportional, and RMSE increases at a slower rate as complexity within a patch increases. 
We conclude that \enki\ shows great promise to surpass 
in-painting as a means of reconstructing cloud masking. 
Future research will develop
\enki\ to reconstruct real-world data. 
\end{abstract}

\section*{Plain Language Summary}
This thesis presents a solution to the problem of cloud masking in the analysis of sea surface temperature (SST) data generated by remote sensing technology. With the vast amount of SST data generated each year, unsupervised machine learning algorithms have proven effective in identifying patterns within the dataset. However, cloud coverage can interfere with the analysis of the data as convolutional neural network models like \ulmo\ are not equipped to handle masked images. To address this issue, we propose an unsupervised machine learning algorithm, called \enki\, which uses a Vision Transformer with Masked Autoencoding to reconstruct pixels that are masked out by clouds.

%% ------------------------------------------------------------------------ %%
%
%  TEXT
%
%% ------------------------------------------------------------------------ %%

%%% Suggested section heads:
% \section{Introduction}
%
% The main text should start with an introduction. Except for short
% manuscripts (such as comments and replies), the text should be divided
% into sections, each with its own heading.

% Headings should be sentence fragments and do not begin with a
% lowercase letter or number. Examples of good headings are:

% \section{Materials and Methods}
% Here is text on Materials and Methods.
%
% \subsection{A descriptive heading about methods}
% More about Methods.
%
% \section{Data} (Or section title might be a descriptive heading about data)
%
% \section{Results} (Or section title might be a descriptive heading about the
% results)
%
% \section{Conclusions}

\section{Introduction}

Remote sensing technology has revolutionized our ability to monitor and understand Earth's systems. Sensors like the Visible Infrared Imaging Radiometer Suite \citep[VIIRS;][]{viirs} or the Moderate Resolution Imaging Spectroradiometer (MODIS; https://oceancolor.gsfc.nasa.gov/data/aqua/) are used for tasks such as weather forecasting, monitoring climate change, atmospheric studies and more. 
One of the most important oceanographic
variables to observe is sea surface temperature (SST), 
which impacts and or tracks key aspects of our ecosystem, e.g., marine dynamics, weather patterns, global warming \citep{sst_examples}.
VIIRS and MODIS both capture SST data at high-spatial resolution ($\approx 1$\,km) with global coverage and twice-daily measurements 
for the past one (VIIRS) or two (MODIS) decades.

Sensors like VIIRS have enabled routine monitoring of the ocean, but because remote sensing technology generates such large datasets (e.g.,\ the National Oceanic and Atmospheric Administration's Level-2P (L2P), 2nd full-mission reanalysis (RAN2) of the VIIRS dataset is nearly  100Tb in size), there is a growing need for automated and efficient ways to sort through and analyze the data. Unsupervised machine learning (\ml) is one approach that has shown great promise in this regard. 
Specifically, by applying unsupervised machine learning to remote sensing data, we can extract meaningful information from large datasets without the need for human intervention or (expensive) labeling. 
This enables one, for example, to observe ocean patterns that may reveal the impact 
of global warming or to identify shifting currents, upwelling or marine
heat waves that impact human activity \citep{ocean_patterns,currents,upwelling,pbg2023}. 
Within Physical Oceanography, such models may identify patterns or outliers in sea surface temperature data, such as the development of oceanic fronts or eddies 
\citep{fronts,eddies}. 

\begin{figure}[ht]
\noindent\includegraphics[width=\textwidth]{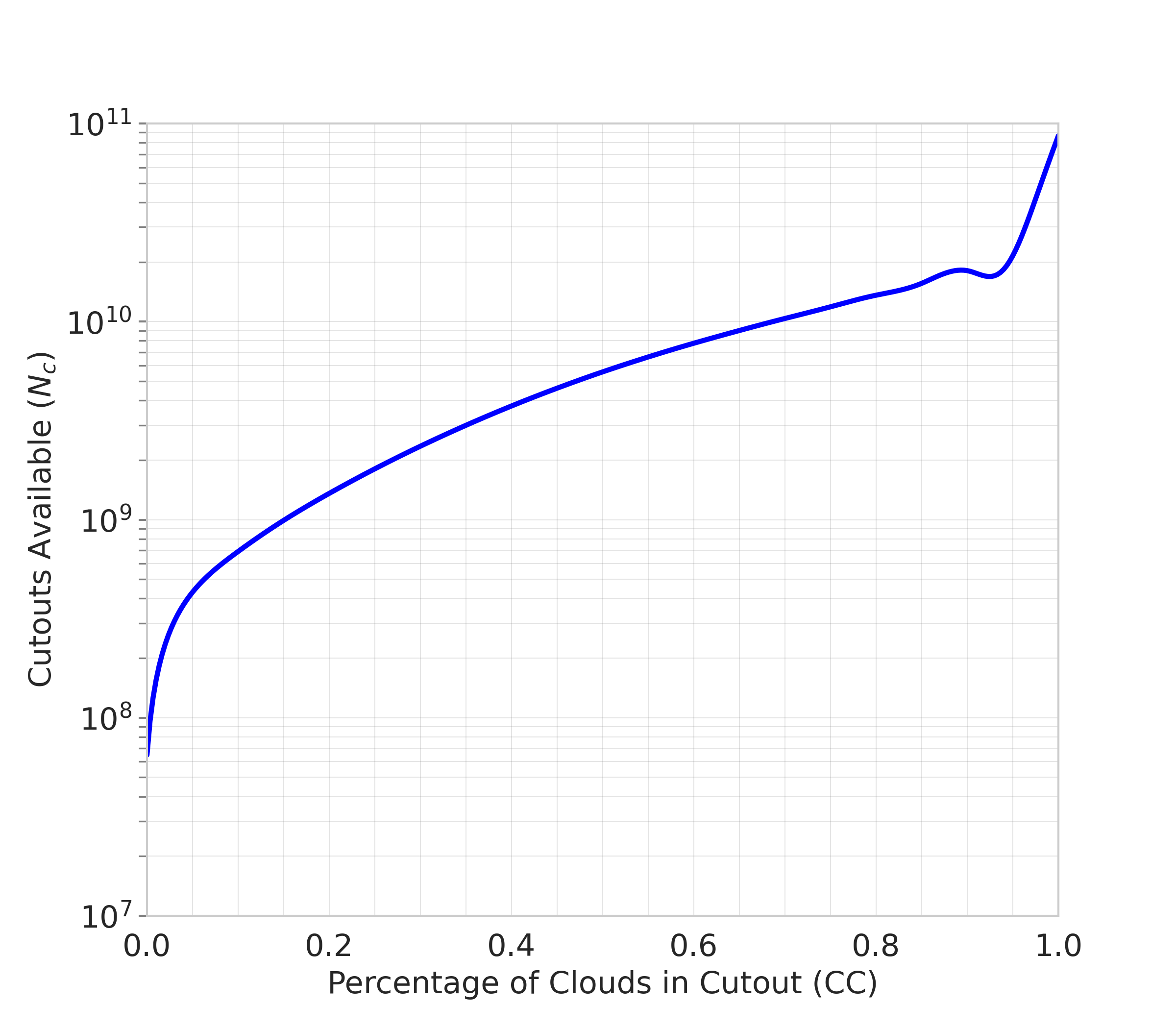}
\caption{Cumulative number of cutouts $N_c$ that are available for
analysis as a function of cloud coverage \CC.
At small values ($\CC < 10\%$), there is a greater than exponential
rise in the amount of data available for analysis.
Increasing from $\CC = 5\%$ to 20\%\ would make $\approx 3\times$
more data available and one would gain an order-of-magnitude
by extending to $\CC = 40\%$.
}
\label{fig:cloud_coverage}
\end{figure}

\begin{figure}[ht]
\noindent\includegraphics[width=\textwidth]{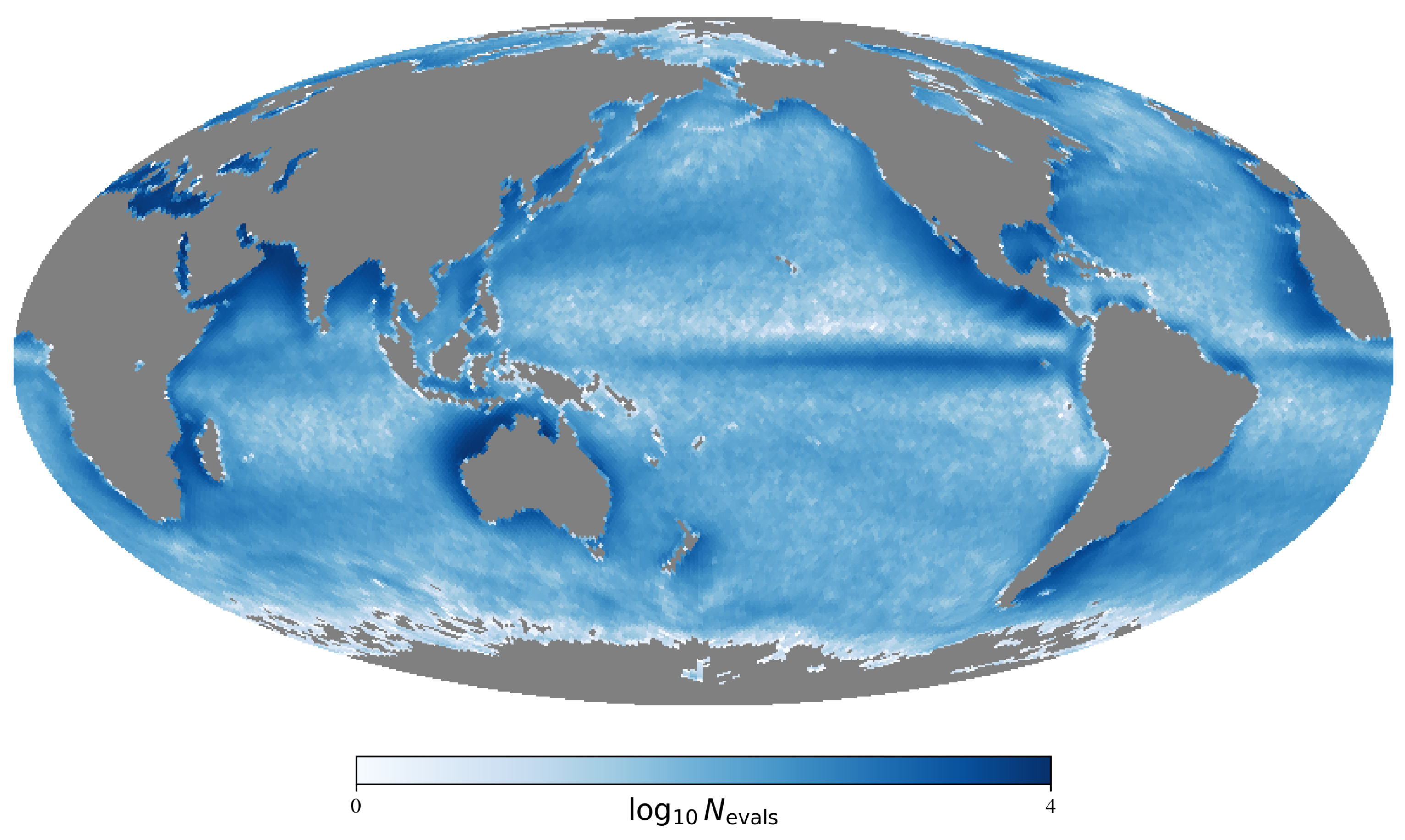}
\caption{
Geographical distribution of the data with $\le 5\%$ masked data (primarily clouds)
in the nighttime L2 MODIS dataset of SST imagery
\citep[see][for full details]{ulmo}.  
Clearly, limiting to nearly cloud-free data greatly limits SST
analysis in various regions.
Note the color bar is on a log10 scale.
}
\label{fig:spatial}
\end{figure}

With the rise of \ml, a type of artificial intelligence (\ai), analysis of 
remote sensing data, a difficult challenge inherent to these data rises to the fore:  corrupt data. 
Clouds, the most impactful corruption, stymie 
remote sensing observations leading to masking and missing data on nearly 
all spatial scales.  These are identified with custom algorithms
and flagged in the Level~2 products produced
by NASA and NOAA among others \citep{cloud_noise}.
Corrupt data can greatly reduce the application and analysis of remote sensing data (Figure~\ref{fig:cloud_coverage}), especially in select geographical regions  (Figures~\ref{fig:spatial}). 
The problem is especially acute for many standard deep-learning \ai\ models 
(e.g.\ convolutional neural networks, CNNs) which generally 
cannot do not allow for missing or masked data.

For example, consider the unsupervised ML model \ulmo\ by our team \citep{ulmo},
a probabilistic autoencoder architecture originally trained on SST images that span 150x150km$^2$ regions, which this paper will refer to as cutouts,
from the L2 SST Aqua MODIS dataset, and later retrained on RAN2 VIIRS SST data \citep{ulmo_on_llc}. 
\ulmo\ utilizes two distinct algorithms: an autoencoder and a normalizing flow. 
The autoencoder reduces the dimensionality of the original image into a latent vector, which represents the model's understanding of the image. These latent vectors are then processed through a normalizing flow to convert them into a single variable known as the log likelihood (\LL). 
The lower the LL, the more uncommon the image
and the greater its complexity \cite{ulmo}. 
This allows one to perform outlier detection to find interesting patterns.

However one major limitation of \ulmo\ and CNNs
in general is that the input images need to be ``whole''. For SST measurements derived from near-infrared wavelengths, therefore, clouds pose a major problem. 
In the case of \ulmo, the authors primarily trained on cutouts that are cloud-free or 
have low cloud coverage ($\CC \le 5\%$), and used a standard inpainting 
algorithm to fill in the masked pixels. 
This resulted in a biased analysis of the ocean with the probability of a cutout decreasing away from the continental boundaries except in the Equatorial region (Figure~\ref{fig:spatial}),
and a significantly reduced fraction of the total data available
(Figure~\ref{fig:cloud_coverage}).

Figure~\ref{fig:cloud_coverage} highlights the potential gains of cloud
mititgation.  Even extending to data from $\CC = 5\%$ to $\CC = 20\%$ would 
increase the dataset by $\sim 3\times$ while extending to 
40\%\ would increase it by over one order of magnitude. 
However, our experiments with the inpainting algorithm 
used in the \ulmo\  \citep[Navier-Stokes][]{bertalmio+2001}
and subsequent analyses indicate poor performance at $\CC > 5\%$, and even reveal important systematic errors at lower cloud coverage fractions.

To make progress on this critical issue, we propose a new solution -- use a novel, ML algorithm based on natural language processing (NLP)
to reconstruct masked data. 
Our model, named \enki, is a Masked Autoencoder built upon the architecture of a Vision Transformer (ViTMAE) trained on LLC SST cutouts \citep{vitmae}. 
By reconstructing the masked out pixels, this allows for SST cutouts to be analyzed by algorithms like \ulmo, and more broadly,
provide accurate estimates for missing data for other scientific
and commercial applications.

This Masters thesis to the Department of Scientific Computing and Applied Maths at the University of California, Santa Cruz describes the development, testing, and first results from \enki. The thesis is organized as follows: Section~\ref{sec:arch} covers the architecture of \vitmae. Section~\ref{sec:method} goes over the environment in which \enki\ was trained, the specifics of our training dataset and validation set, and the hyperparameters we used during training. Section~\ref{sec:results} goes over the qualitative and quantitative results as we examine \enki\ 's reconstructions.

\begin{figure}[ht]
\noindent\includegraphics[width=\textwidth]{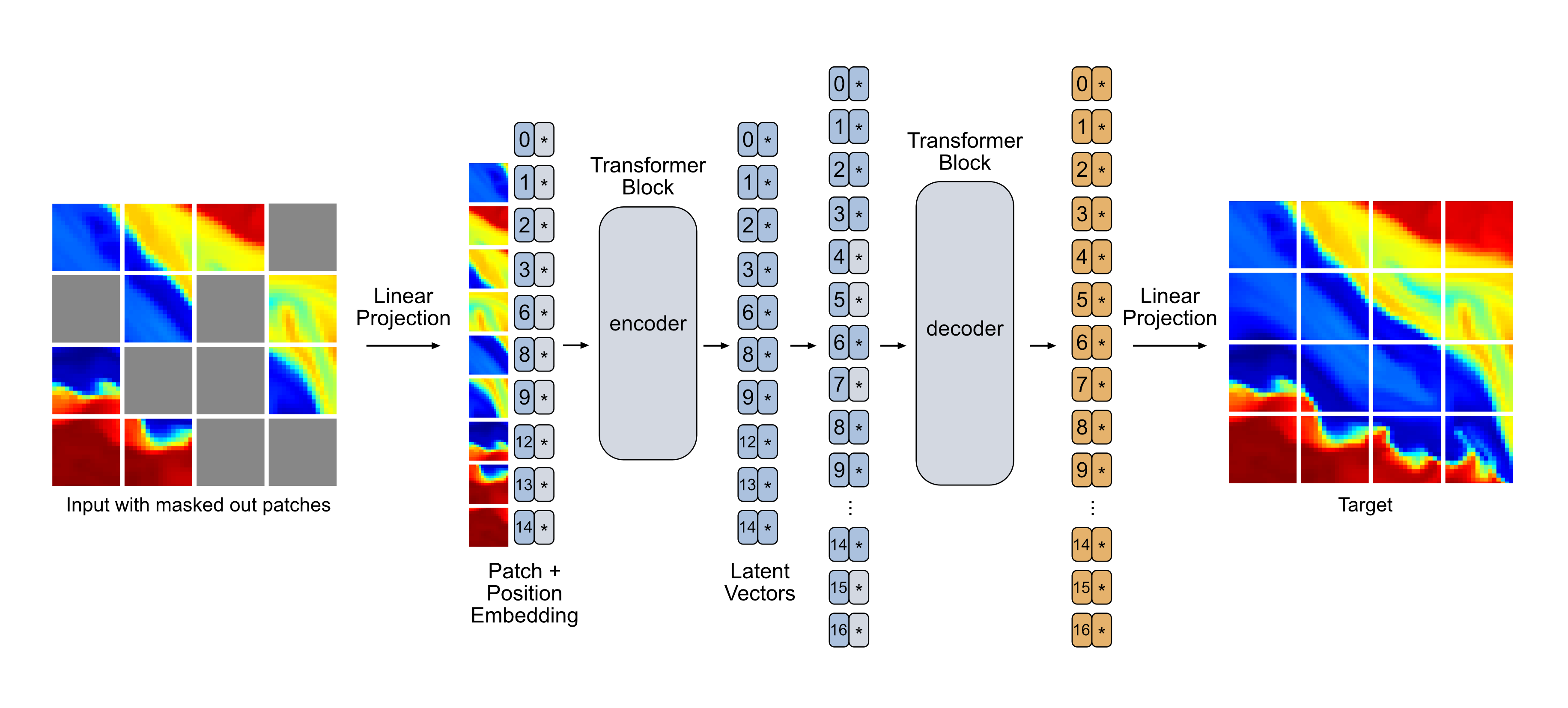}
\caption{Architecture of our ViTMAE named \enki. In this example, the image is 64x64 pixels and broken up into patch sizes of 16x16 pixels with ~40\% of the image masked out.
An image is broken down into patches and masked. The unmasked patches are flattened and embedded by a linear projection with positional embeddings which are then run through the encoder, returning the encoded patches which are then run through the decoder along with the masked (unfilled) tokens. 
This returns another latent vector, and another linear projection layer outputs this vector as an image with the same dimensions of the original image. 
The model of \enki\ we train reconstructs cutouts of size $64\times64$\,pixel$^2$ with a patch size of $4\times4$\,pixel$^2$ and a latent vector size of 256.}
\label{fig:arch}
\end{figure}

\section{Architecture}
\label{sec:arch}

\enki\ is a Masked Autoencoder (MAE) neural network with 
a Vision Transformer (ViT) architecture \citep{vitmae}. 
As an autoencoder, it has two primary parts: (1) the encoder and (2) the decoder \citep[e.g.][]{baldi2012autoencoders}. The encoder specializes in transforming an input image into a lower-dimensional latent space, known as a latent vector. In machine learning, latent vectors are reduced, numerical representations of data which (ideally) capture the essential characteristics or features of the data. The decoder transforms this latent vector back into a reconstructed output image with the same dimensions of the input space. 

Transformers are another type of neural network that also have an encoder and decoder type architecture but differ from autoencoders in their structure and use. Unlike autoencoders, transformers tokenize inputs. Each token contains a latent representation, and the transformer applies self-attention to weigh the contribution of each latent vector to the final data representation. Self-attention works by computing a set of attention weights for each latent vector, based on its similarity and association
to other latent vectors in the image. Transformers are primarily used in NLP for tasks such as language translation. An example would be if a transformer was given an English sentence with the goal of reconstructing the sentence in Japanese. The transformer tokenizes each word, performs self-attention to find the relations between the words and produces the latent vector, and then reconstructs the sentence 
in Japanese from the latent vector.

A ViT is a Transformer encoder applied to images \citep{vit}. Similar to a sentence, the image is tokenized by breaking it into non-overlapping patches, each of which is assigned a positional embedding. These are then fed through a standard Transformer encoder. Self-attention is also performed on the patches to weigh contributions of each patch to the final image representation. ViT is primarily used for image classification (i.e.,\ without a Transfomer decoder) and is therefore a supervised learning model requiring labeled data for training.

\cite{vitmae} developed a masked autoencoder based on the ViT which utilizes both the encoder and decoder and learns to reconstruct images with masked/missing data. The masked nature allows for unsupervised training which is ideal for handling unlabeled big data. \vitmae\ works by taking an image, breaking it up into patches, masking out a portion of the patches, and then tokenizing them. These patches are then fed through a transformer (the encoder), and a latent vector is returned. The latent vector is then given to another transformer block (the decoder), which reconstructs the image and a linear projection layer converts it back into the original image's dimension size.

Our architecture, named \enki\ and described in Figure~\ref{fig:arch}, follows closely that of the original \vitmae. We changed the hyper parameters of image size to fit our $64\times64$\,pixel$^2$ cutouts, and reduced the patch size down to $4\times4$\,pixel$^2$. 
A small patch size is preferred because real clouds are not large squares. 
A future version of \enki\ 
will convert cloud mask into a set of patches, and
we can reproduce a mask with squares while minimizing
the number of good pixels masked.

On the other hand, smaller
patch sizes are more computationally expensive. 
Decreasing the patch size down to even $2\times2$\,pixel$^2$ will increase the total 
number of patches from 256 to 1024 with an increase of $4\times$ in compute. 

\begin{table}[]
\centering
\begin{tabular}{|l|c|}
\hline
  Config & Value \\ 
  \hline
%\textbf{t10} & 1.018e-0.6 & 1.214e-06 & 8.741e-0.6 & 8.027e-05 & 9.601e-05 \\ \hline
Models and Masking Ratios & t=10, 35, 50, 75 \\
Base Learning rate (t=10, 35, 50) & 1e-4 \\
Base Learning rate (t=75) & 1.5e-4 \\
Weight Decay & 0.05 \\
Warmup epochs & 40 \\
Training Epochs  & 400 \\
Batch size &64 \\
Image size &(64, 64, 1) \\
Patch size &(4, 4) \\
Encoder Dimension & 258 \\
Decoder Dimension & 512 \\
  \hline
\end{tabular}
\caption{The training hyperparameters for the model are summarized in this table, highlighting the different learning rates chosen for each specific model. Additionally, a 256 embedding dimension was initially selected with the intention of utilizing the latent vectors for training \ulmo. However, it was discovered that this approach was not feasible due to the computationally expensive nature of training \ulmo\ with each patch generating a 256-sized latent vector. As a result, alternative strategies were explored to address this limitation.}
\label{tab:hyper}
\end{table}

\begin{figure}[ht]
\noindent\includegraphics[width=\textwidth]{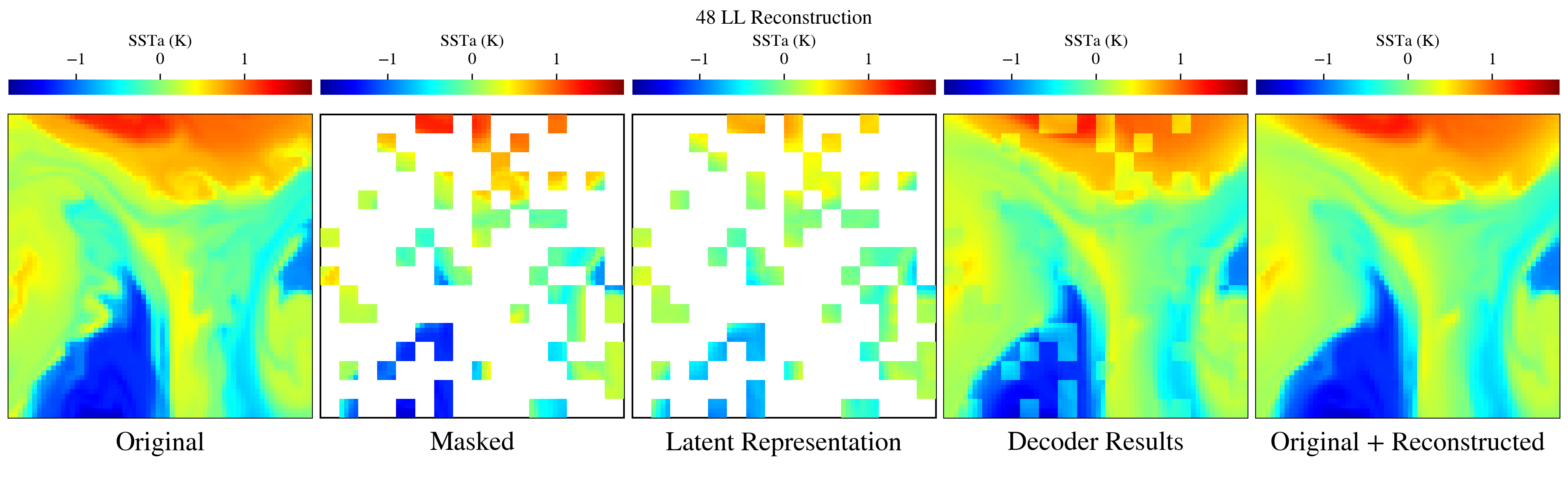}
\caption{A visual example of the reconstruction process. 
1) We start with the Original image; 
2) We convert this image into patches, and mask out a \pper\ percentage of the image. In this example, we use $\pper=75$;
3) The unmasked patches are run through a transformer block (the encoder; see Figure~\ref{fig:arch}) 
that returns latent representations of the patch. Each unmasked patch has a latent vector of size 256;
4) The full set of patches (latent vectors of unmasked patches + empty masked patches) are now run through another transformer block (the decoder) that reconstructs the missing patches. This image is converted back into the original dimension space of the original image. However, notice that what the decoder returns has notable patches where the image was not masked. We suspect this is because when the model is evaluating MSE, it only evaluates the reconstructed masked pixels. This means when reconstructing, the model does not care what the unmasked patches look like; 
and
5) We take the unmasked patches from the Original image and replace the missing patches with the reconstructed patches from the decoder to create our final reconstructed image.}
\label{fig:recon_example}
\end{figure}

\section{Methodology}
\label{sec:method}

\enki\ is trained in two steps: pre-training and fine-tuning. Pre-training is typically used to train 
a deep-learning model on a general corpus, 
while fine-tuning is used to adapt the model to a specific dataset. A model of \vitmae\ could be trained on a general corpus like ImageNet-1K and then fine-tuned on another dataset to specialize in reconstructing birds. However, when a model of \vitmae\ is trained, the hyperparameters used during training cannot be changed when reconstructing new images outside of training. 
For example, 
a model of \vitmae\ was trained on 128x128, three color channel images 
with a 16x16 patch size,  must be applied to data with the same shape.
Our training set did not match the parameters of the available trained models. 
Therefore \enki\ was pre-trained from scratch to suit our needs. Pre-training for \enki\ was performed using a modified version of Facebook AI Research group’s code to run in Kubernetes with an h5 file containing our training dataset. Multiple models of \enki\ were trained on different masking ratios (\tper) to study 
how its affect on \enki\ performance for image reconstruction.

Finally, we have experimented with different masking ratios during and after pre-training. We distinguish these two values as the masking ratio a model was trained at as \tper\ 
(e.g. a model trained at 10\% masking is referred to as \tper=10) 
and the masking ratio for the input image 
for reconstruction is referred to as \pper\ 
(e.g. an image being reconstructed at 10\%\ masking is referred to as \pper=10).

\subsection{Training Dataset}
\label{sec:training_dataset}

In unsupervised learning, the dataset does not have labels. 
Our training dataset is comprised of millions of SST images
for \enki\ to learn patterns. 
For this project, the optimal dataset would be one that is free of clouds and noise because we do not want \enki\ to learn to reconstruct unmasked clouds or add noise into reconstructed patches\citep{cloud_noise}. The dataset should adequately capture both typical and uncommon SST dynamics with a uniform coverage of the ocean.

For pretraining, we used a dataset generated from 
a Ocean General Circulation Model (OGCM) from the
Estimating the Circulation and Climate of the
Ocean (ECCO) project in a collaborative effort between the 
Massachusetts Institute of Technology (MIT), the Jet Propulsion
Laboratory (JPL), and the NASA Ames Research Center (ARC).
Specifically, we use the $\frac{1}{48}^\circ$, 90-level simulation known as LLC4320. 
The \llc\ model outputs do not contain clouds (only a statistical model of the atmosphere) 
or sensor noise. 

Furthermore, while we have large sets of cutouts of true data from MODIS and VIIRS that are nominally “cloud free”, our experience is that many of these contain clouds that were missed by standard screening algorithms \citep[see][for examples]{ulmo_on_llc}. 
Working with a cloud-free VIIRS dataset may also bias \enki\ towards behaving
best in cloud-free regions (e.g. Figure~\ref{fig:spatial}). 
\llc\ data does not suffer from this issue; we can generate a training 
set with uniform coverage of the ocean in both location and time. 
Additionally, \cite{ulmo_on_llc} shows that the distribution of sea surface 
temperature anomaly (\ssta) patterns present in the VIIRS observations are generally well-predicted by the LLC model. 
This provides confidence that a \vitmae\ trained on \llc\ outputs
may well reflect patterns within the real ocean.

The training dataset for \enki\ consists of \llc\ cutouts from  
2011-11-17 to 2012-11-15, inclusive. 
\llc\ cutouts were extracted to cover $144\times144$\,km$^2$ sampling and resized
using the local mean to $64\times64$\,pixel$^2$ cutouts. 
The geographical distribution uniformly samples the ocean between 
latitudes of 90S and 57N.
Real satellite data contain noise from the atmosphere and from instruments, but because the goal of reconstruction is not to reconstruct the noise in images, we did not impute noise
in the \llc\ cutouts \cite[in contrast to][]{ulmo_on_llc}. In total, the training dataset ends up being $\sim 2.6$~million cutouts. 
Although it is possible to generate larger datasets for training purposes, we encountered resource limitations during the training process, and expanding the dataset size would significantly extend the training duration. 
Unlike datasets such as ImageNet, which contain a diverse range of objects like animals, plants, and various objects, SST patterns exhibit less variability. 
Given this characteristic, we hoped that the inclusion of 2.6 million cutouts in our dataset adequately captures the dynamics present in the ocean.

\begin{figure}[ht]
\noindent\includegraphics[width=\textwidth]{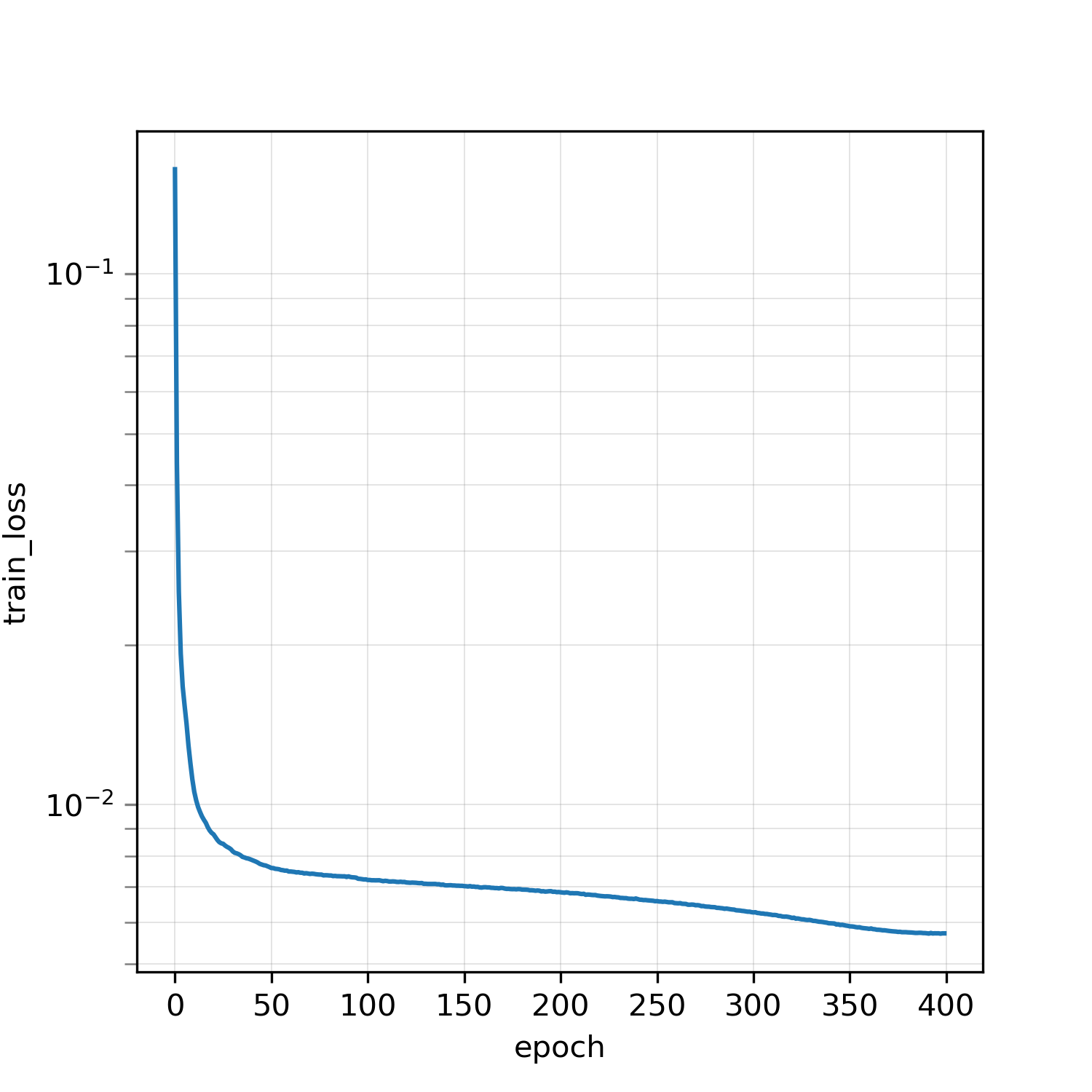}
\caption{Training loss of the $\tper=75$ \enki\ model. 
The lost function implemented was
the MSE of the reconstructed masked out patches.
As typical of deep-learning models, there is rapid improvement
in the loss at early stages followed by a shallow plateau at late times.
}
\label{fig:loss}
\end{figure}

\subsection{Training}
\label{sec:training}

We performed a hyperparameter scan of the masking ratio during training \tper. 
Four different models of \enki\ were trained with 10\%, 35\%, 50\%, and 75\% masking. 
These variations are referred to as $\tper=10,35,50,75$ respectively. 
Each model was trained for 400 epochs on the complete dataset of 2.6 million \llc\ SST cutouts. 
During training, several models errored if the learning rate was too high. 
For these models, 
we adjusted the learning rates from the default value;
the $\tper=75$ model was trained 
at 0.00015, while the $\tper=10,35,50$ models were trained at 0.0001.
During a single training epoch, \enki\ iterates through every cutout 
in the dataset and masks out the specified masking ratio with random 
$4\times4 \, \rm pixel^2$ patches. 
Performance was evaluated using the default Mean Squared Error (MSE)
loss function, but going forward we may consider implementing other loss functions. 

Every epoch of training uses new randomized masks for each cutout. 
This forces the model to see significant variations and better generalize.
All models were trained in the Nautilus Pacific Rim Platform
with Kubernetes using 4 CPUs, 75Gb of RAM, and 8 NVIDIA-A10 GPUs. 
These specs were chosen after profiling training speeds with available resources. Training times varied based on masking ratios. 
Higher masking ratios completed in shorter training times, 
with the $\tper=75$ model training the fastest ($\approx 22$ minutes per epoch).
Lower masking ratios require more computations to create the latent vectors; 
the $\tper=10$ has a 40\% longer training time ($\approx 30$ minutes per epoch).

\subsection{Validation Dataset}
\label{sec:valid}

In unsupervised learning, the validation set is a subset of the data not used during training.
It is used to evaluate the performance of a model with data the model has not yet seen. 
The validation set therefore assesses model performance while avoiding any effects
of overfitting during training. Overfitting refers to when a model learns the dataset too well that it is able to give accurate predictions, or in our case reconstructions, for the training dataset, but has not generalized well to reconstruct new data. 
While 2.6 million cutouts is a substantial dataset, 
it is still small compared to the 100Tb of data in the VIIRS dataset. If we want \enki\ to perform well, 
we wish it to reconstruct data outside of the training set. 

The validation set we created for this project was a batch of 680,000 \llc\ cutouts 
from the same locations of the training dataset but offset by 
$\approx 2$~weeks from the training set. 
These cutouts follow similar cloud free and no noise and 
were drawn uniformly over the ocean. 
While not within the scope of this thesis, the validation set for future studies could contain VIIRS or MODIS cutouts to determine \enki's effectiveness reconstructing real cutouts.

% %%%%%%%%%%%%%%%%%%%%%%%%%%%%%%%%%%%%%%%%%%%%%%%%%%%%%%%%%%%%%%%%%%%%%%%%%%%%%%%%%%%%%%
\section{Analysis and Results}
\label{sec:results}

Following the completion of training, we used the validation set to reconstruct cutouts with the four \enki\ models trained at $\tper=10,35,50,75$. For each model, we reconstructed images at a masking ratio of $\pper=10,20,30,40,50$. 
We chose to limit our range to $\pper \le 50$ because further masking would run 
the risk of \enki\ having insufficient information to interpolate accurately. 
The seed for the masking was not set, so masks for all the reconstructions differ.
Reconstructed images were created by replacing masked patches with reconstructed patches.
The results are invariably imperfect and we anticipate larger
errors at higher masking ratios. 
To assess the magnitude of the error, we calculate
the Root Mean Squared Error (RMSE):
\begin{equation}
    {RMSE}(y, \hat{y}) = \sqrt{\frac{\sum_{i=0}^{N - 1} (y_i - \hat{y}_i)^2}{N}}
\end{equation}
where $y$ represents the original image, $\hat{y}$ represents the reconstructed image, and $N$ represents the number of pixels that were masked out. Throughout, the RMSE is only calculated on masked pixels $i$.

Analysis of the RMSE is performed at the patch level and on images as a whole (excluding the edge patches, which we will discuss in depth in Section~\ref{sec:patch_recon}). 
As discussed in Section~\ref{sec:bias}, we identified a bias for select \tper,\pper\ values.
In nearly all cases, we correct for this bias before evaluating the RMSE.
In the following sub-sections, we present results including
the performance of the \enki\ models with various \tper\ values. 

\begin{figure}[ht]
\noindent\includegraphics[width=\textwidth]{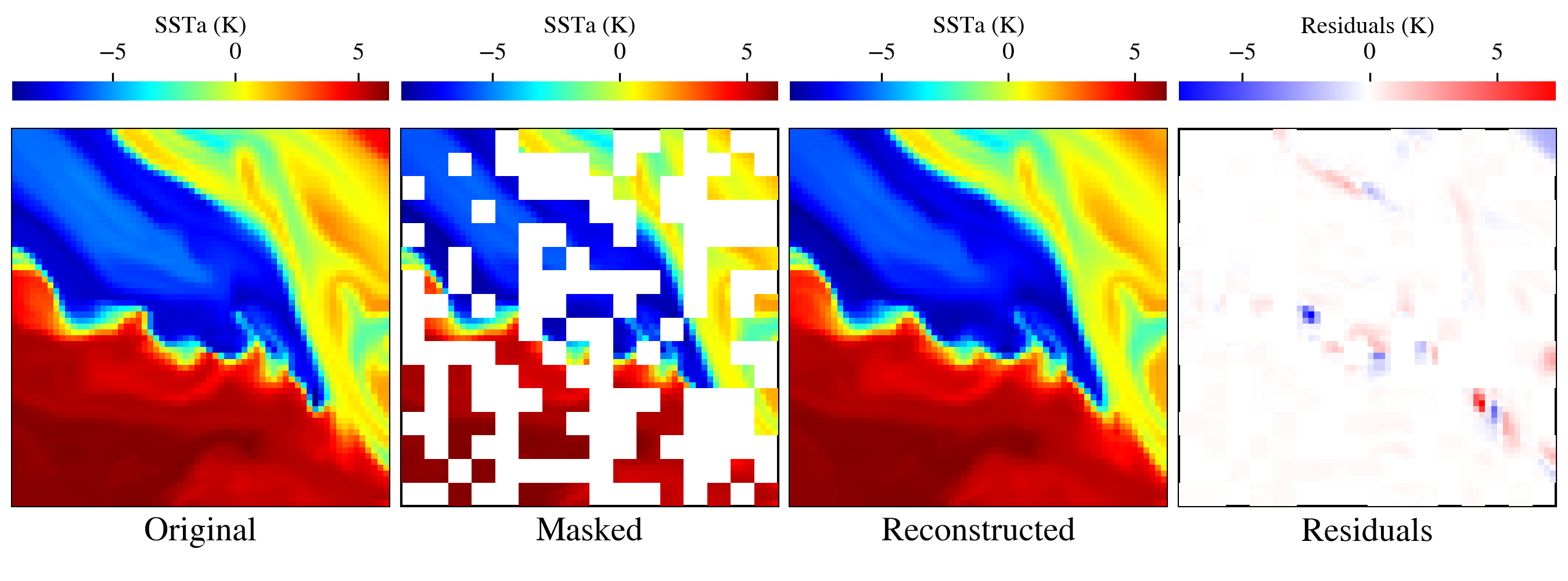}
\caption{
Qualitative assessment of the reconstruction
for a single image with \enki.  This high-complexity
image ($\LL \approx -19000$) was imputed with
a $\pper=50$ mask and we applied the $\tper=50$
\enki\ model.
Despite the high masking ratio, the reconstruction
reproduces the original at high fidelity.
The primary exception are features along the 
sharpest gradient (residuals of 7.3 and -7.5) and
the upper-right corner (residuals of -2.5) where the model is forced
to extrapolate.
}
\label{fig:corner_example}
\end{figure}

\subsection{Qualitative Results}
\label{sec:decoder}

\begin{figure}[ht]
\noindent\includegraphics[height=0.8\textheight]{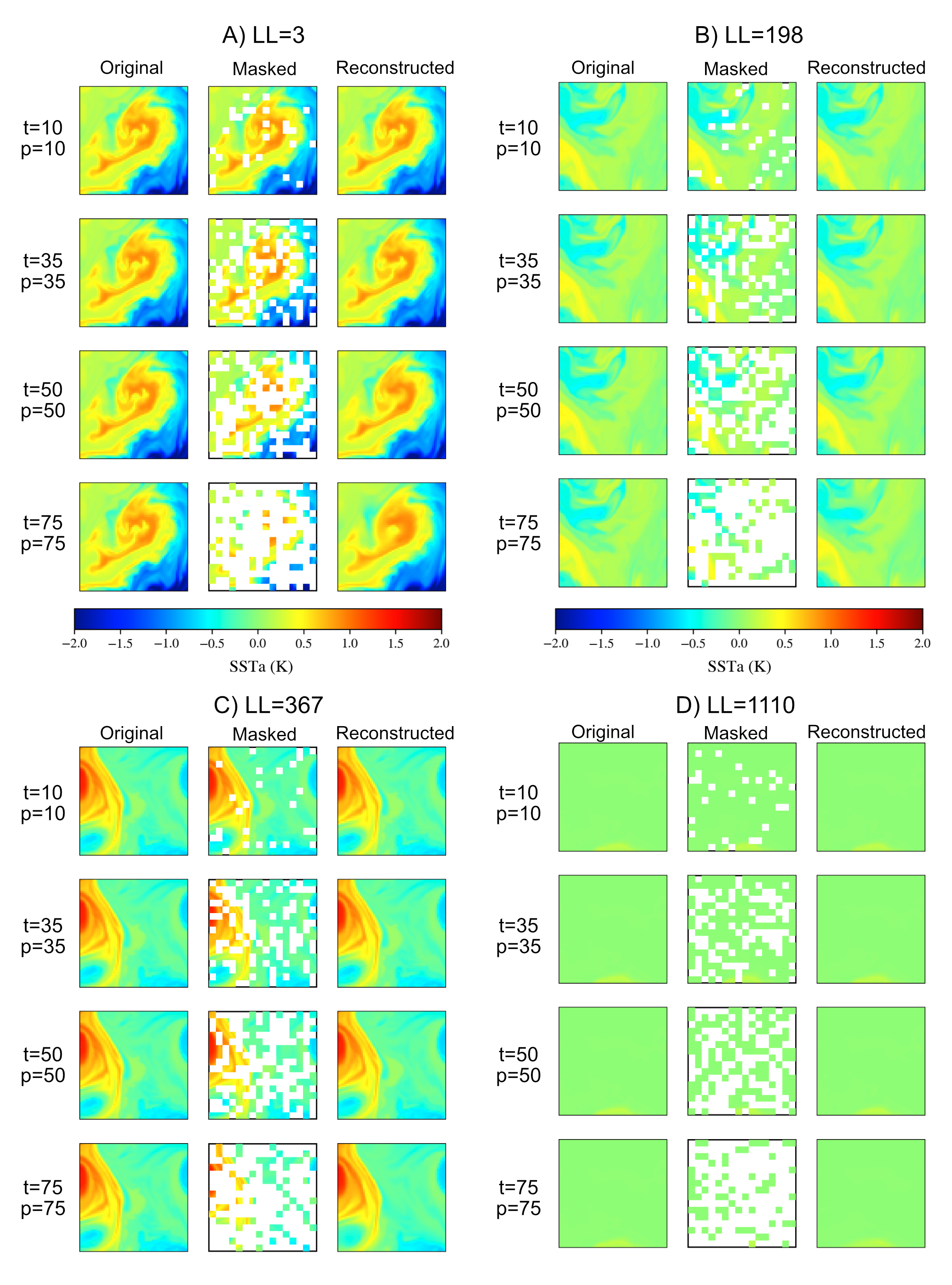}
\caption{
Galleries of reconstructions by the different models. 
In each case, the cutout is 
reconstructed using the \enki-trained model that matches the masking ratio \pper. 
A collection of reconstruction galleries showcasing the results obtained from various models is presented in this study. Each gallery exhibits the reconstruction of a different cutout being reconstructed by a model at the mask training ratio they were trained at. Different LL were intentionally selected to demonstrate how \enki\ performs across a spectrum of ocean dynamics.
}
\label{fig:recon_gallery}
\end{figure}

Consider first a visual inspection and assessment of the image 
reconstructions by \enki.
Figure~\ref{fig:corner_example} shows a representative example for a 
higher complexity SST cutout.  
The example has a $\pper=50$ mask ratio and 
we implemented the $\tper=50$ \enki\ model.

Overall, the \enki\ model demonstrates strong performance in capturing the underlying structure of the SST patterns, even at higher masking ratios. 
The reconstructed images exhibit smooth transitions between the unmasked and reconstructed patches and no significant offsets, eliminating the need for bias correction. 
However, some residual errors are observed along sharper gradients, where the model accurately estimates the border between different values but falls short of perfection. 
We also see that in cases where certain features, such as the curved point of negative (blue) values, are completely masked out, \enki\ produces sensible, albeit imperfect, reconstructions. 
A similar occurrence can be observed in the upper-right corner where the gradient is mostly masked out, and the reconstructed gradient is less harsh.

Expanding on the qualitative inspection,
Figure~\ref{fig:recon_gallery} shows the
reconstructions for four images chosen to 
span a range of complexity (here we use the \LL\ metric from \ulmo).
Overall, the reconstructions are close to the original images. 
Reflecting what we saw from Figure~\ref{fig:corner_example}, as the masking ratio increases,
the model begins to miss or incorrectly guess 
some of the features. For example, for image A and B
for $\tper=75, \pper=75$ there are features which
are smoothed over or incorrectly reconstructed. In image C for $\tper=75, \pper=75$ in the bottom right corner there is a small negative value (blue) feature that is entirely missing in the reconstruction. 
When looking at the mask, we find that the feature had been entirely masked out leaving \enki\ no information to interpolate the feature. 
Despite this, \enki\ we are still able to recognize the key patterns from the original image in the reconstruction indicating that \enki\ is able to capture the overall structure of the cutouts. 
We continue to observe these types of incorrect reconstructions in the $\tper=50$, $\pper=50$ examples, but to a lesser extent. In A we see for $\tper=50$, $\pper=5$ that the dot of orange to the mid-left of the image is missing in the reconstruction, but when looking at the blue gradients on the bottom-right and the swirl in the middle, we find the features are not as smoothed out as they are in the $\tper=75$, $\pper=75$ reconstructions. 

Meanwhile, the images from D with LL=1110 are reconstructed well across all models, and we suspect that complexity of a cutout will play 
a role in how well \enki\ reconstructs images.

\begin{figure}[ht]
\noindent\includegraphics[width=\textwidth]{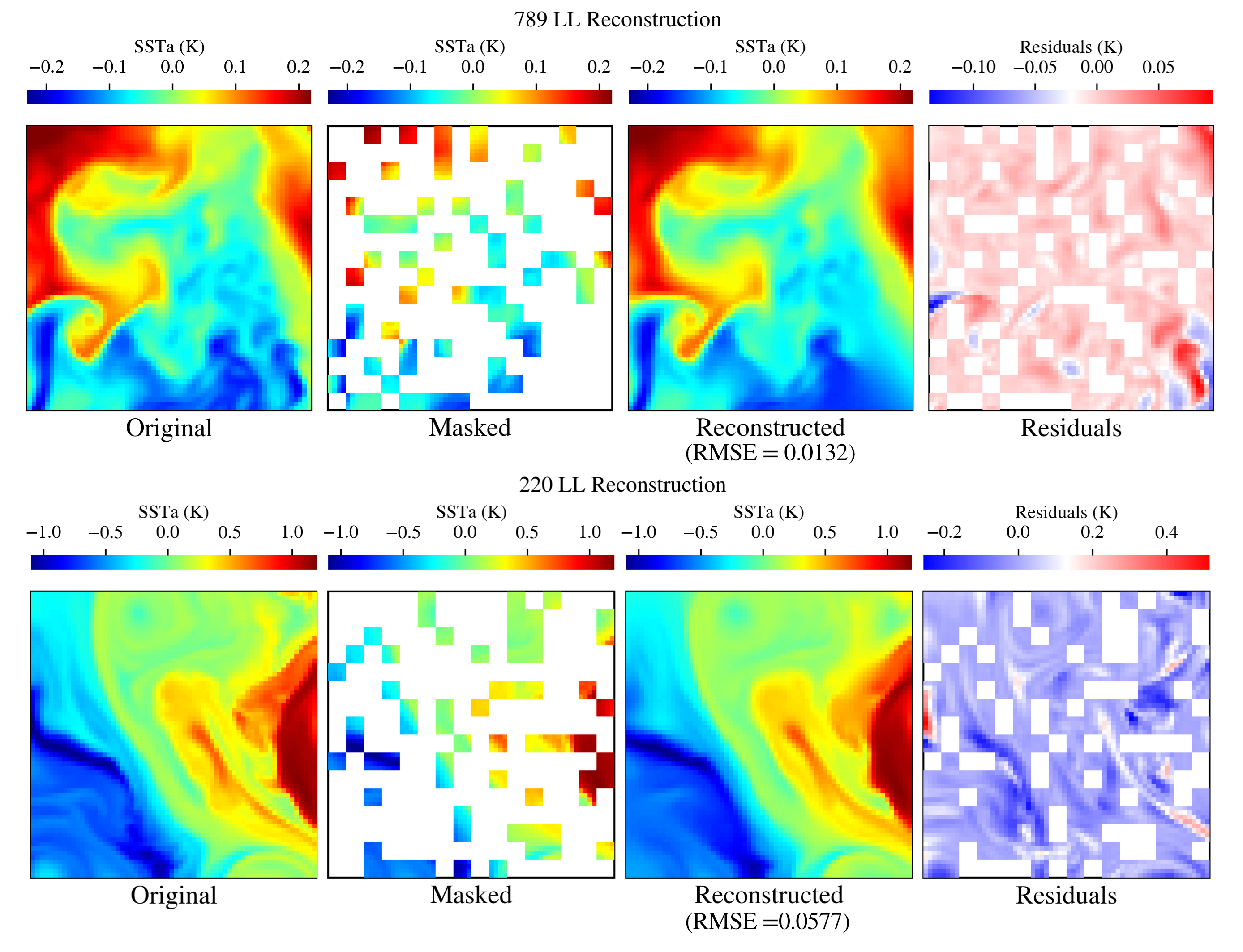}
\caption{ The top image demonstrates how the \enki\ model trained at $\tper=75$ will frequently miss finer features in $\pper=75$ reconstructions. 
Notice how the bottom right corner is almost entirely masked out. In the reconstruction, all the structure of the original cutout is replaced with a smooth gradient. 
The bottom image demonstrates how the $\tper=75$ model
generalizes shapes, but looses a lot of the finer details.
}
\label{fig:t75_p75_recon}
\end{figure}

\begin{figure}[ht]
\noindent\includegraphics[width=\textwidth]{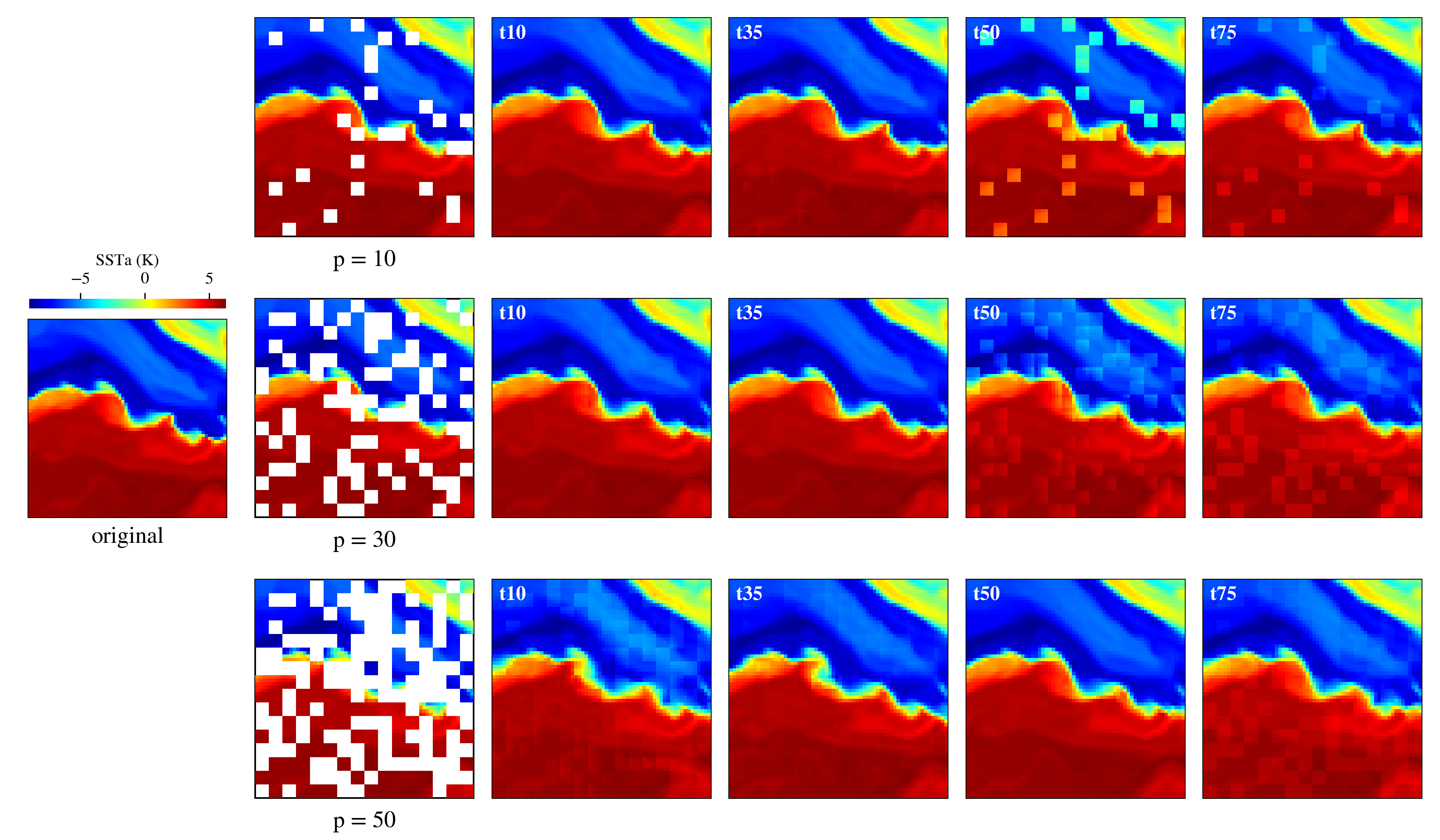}
\caption{Gallery of different models reconstructing cutouts at $\pper=10,30,50$. Notice that a model performs best when reconstructing around the range for which it was trained. Models are still able to capture the values of an image when reconstructing around ~15-25\% away from where they were trained, but the patches become visible to the human eye. This is especially noticeable in the $\tper=35$, $\pper=10,50$ reconstructions.
}
\label{fig:model_comparison}
\end{figure}

We next examined at how models reconstruct at masking ratios
($p$) outside of their training ($t$). In Figure \ref{fig:model_comparison} we show an example of the $\tper=10,35,50,75$ models reconstructing at $\pper=10,30,50$. Here, we see issues arise with the reconstructions. For the $\tper=10$ model, we see erros in the patches at $\pper=30$, and at $\pper=50$ this effect is magnified.
The $\tper=35$ model has this issue for $\pper=10,50$, while the $\pper=30$ reconstruction does not. The $\tper=50$ model performs significantly worse than the rest of the models at $\pper=10,30$, but does well at $\pper=50$. This implies that there may be a certain offset between $t$ and $p$ 
for which a particular model of \enki\ will perform well.

Finally, the $\tper=75$ model does not have a single masking ratio within the chosen range that produces acceptable results. At all levels of \pper\ the patches are visible, and there is
an offset in values from the original image.

In their original paper, \cite{vitmae} found that 
training at 75\%\ masking was ideal because it 
reduced redundancy when training, and forced the model to learn more rather than extrapolating information from nearby patches. 
However, for our purposes, we are more interested in accurate reconstructions over well generalized ones. 
We have already seen that when a lot of smaller, finer features are fully masked out, these features tend to either be smoothed over or missing within the reconstructions. 
We demonstrate more in-depth examples of this in 
Figure~\ref{fig:t75_p75_recon}.  In the top right corner of the
upper image, \enki\ fails to reconstruct the small band, and completely smooths over all the finer dynamics in the bottom right. 
The bottom figure demonstrates another example of \enki\ smoothing out finer details. 

In addition, it was noted in the original MAE paper that different masking techniques were tested. One of these techniques was a “block-wise” masking strategy that masked out a large block of patches from the middle of the image. While random patching is convenient for training due to the high variability in masks, random patching does not accurately represent cloud masks. This is especially true for the $\pper=10$ reconstructions where we have many isolated patches scattered around the cutout. Block masking could more accurately emulate large clusters of clouds blocking out large portions of the image. 
We will consider this approach in future work.

\begin{figure}[ht]
\noindent\includegraphics[width=\textwidth]{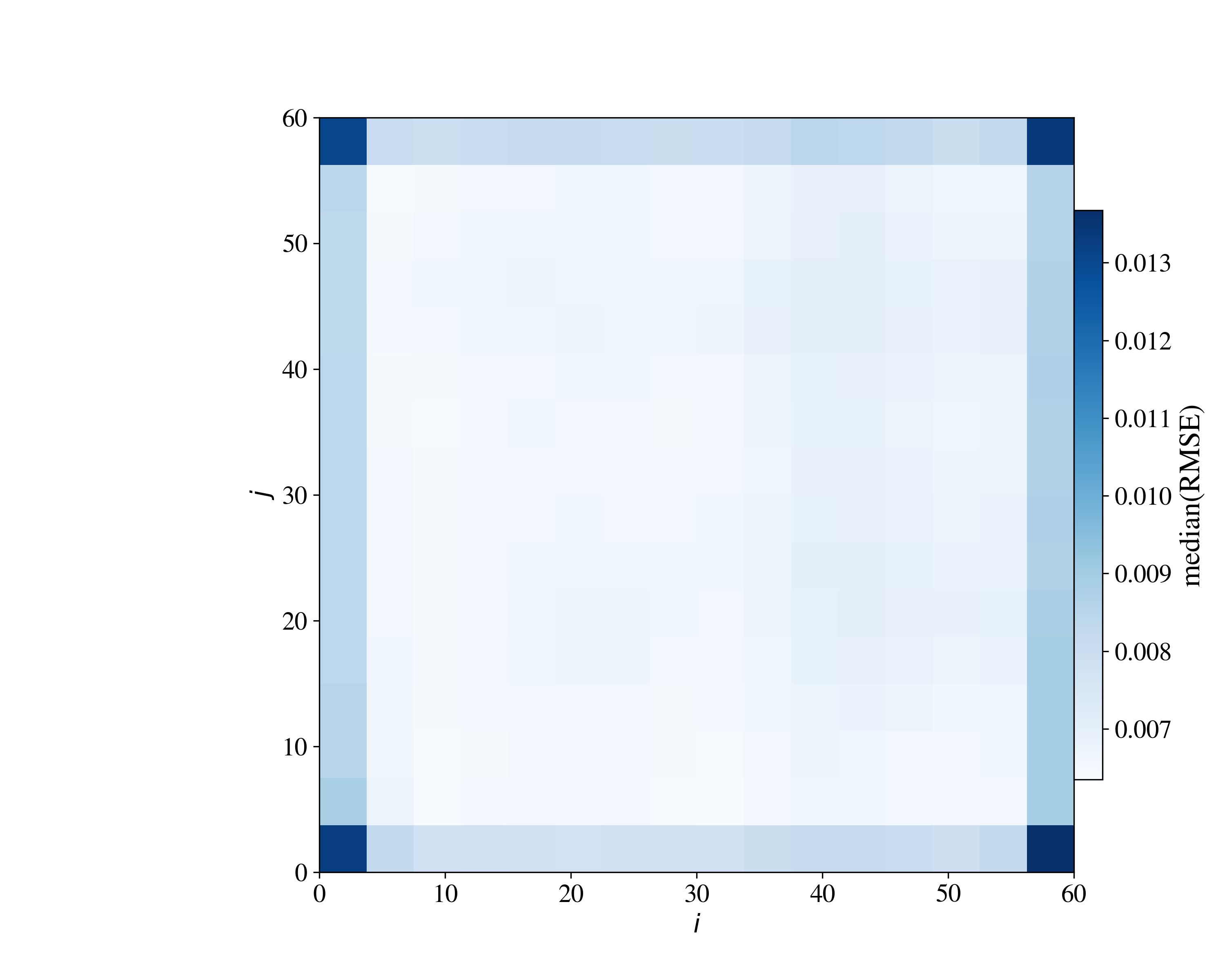}
% TODO
% 2. Remake with t=10, p=10 or p=20.  The current is t=75,p=20
\caption{
Average RMSE error for reconstruction of all masked 
$4\times4 \, \rm pixel^2$ patches at the location $i,j$ in the image. 
The $i$ and $j$ represent the position of the patch from the lower left corner. 
These are for the validation set with  
$\tper=10$ and $\pper=20$ reconstructions. 
%There is a notable border along the edge of the 
%figure indicating that 
Clearly, patches at the edge suffer
significantly higher RMSE values.  
We attribute this to the fact
that edges and corners have less surrounding information 
compared to patches in the center.
For a real-world application of \enki, we 
would advise ignoring the image-border for any
reconstructions.
All further calculations and analysis presented in this thesis
explicitly ignores the border.
}
\label{fig:patch_visual}
\end{figure}

% %%%%%%%%%%%%%%%%%%%%%%%%%%%%%%%%%%%%%%%%%%%%%%%%%%%%%%%%%%%%%%%%%%%%%%%%%%
\subsection{Analysis of Reconstructions at the Individual Patch Level}
\label{sec:patch_recon}

\begin{figure}[ht]
\noindent\includegraphics[width=0.9\textwidth]{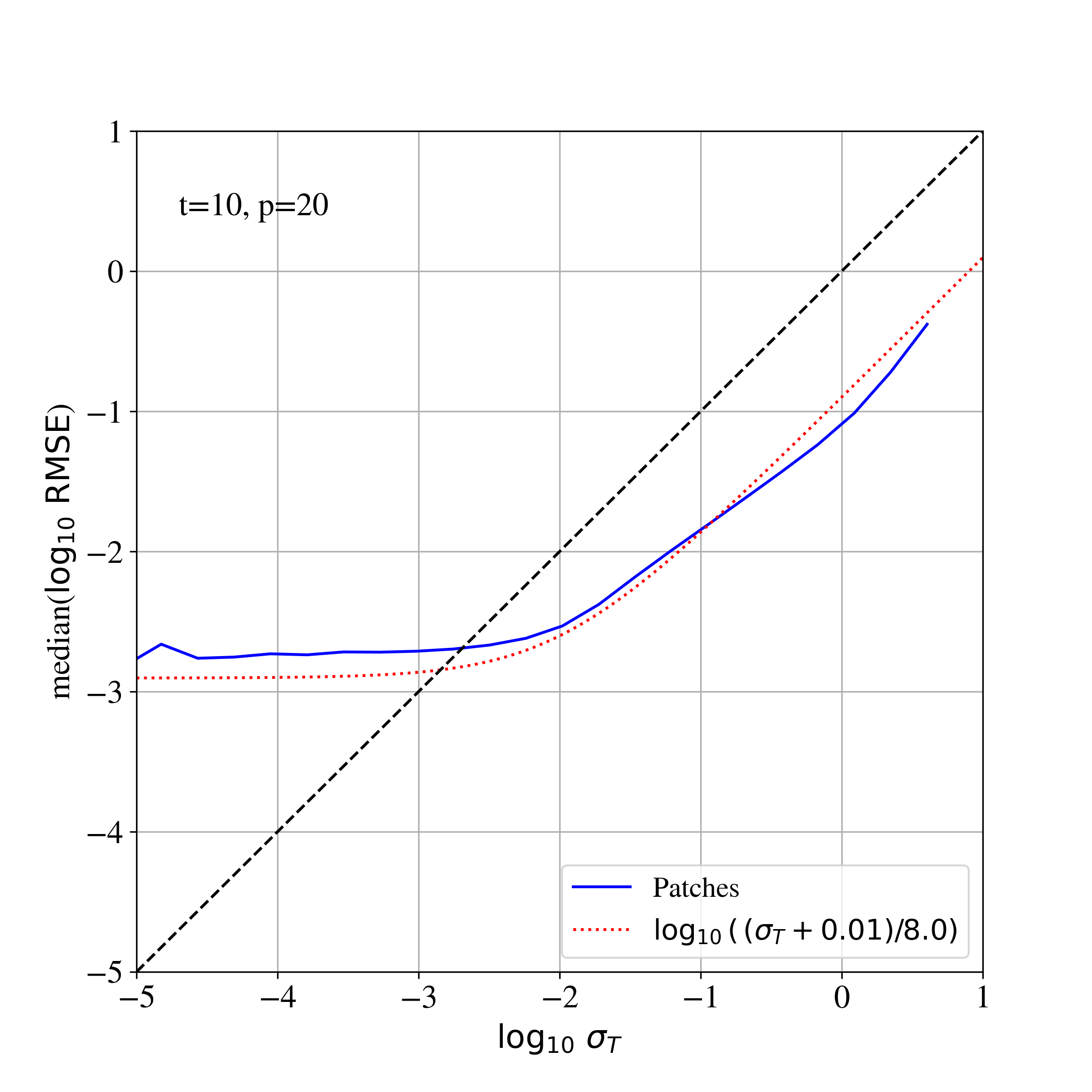}
\caption{
RMSE of patches vs the standard deviation of \ssta (\stdT). 
The black line is the one-to-one relationship where RMSE = \stdT. 
The blue line is the RMSE of $\tper=10$, $\pper=20$ patches vs \stdT, and the red line represents a (by-eye) fit to the blue line. 
We see that for $\log_{10} \stdT >-2$ (0.02K), 
\enki's RMSE is nearly a magnitude less than the \stdT. 
This means the reconstructions are far more accurate than
random noise. 
At $\log_{10} \stdT < -2$, the RMSE flatten out. 
We speculate that this is because spatial scale of the features become too small for outside patches to assist in reconstruction.
}
\label{fig:patch_complexity}
\end{figure}

An analysis of \enki's performance was conducted at 
the individual patch level to evaluate whether patch complexity or position affected \enki's patch reconstruction. 
We begin by calculating the RMSE of every patch of the reconstructed $\tper=10$, $\pper=20$ images, 
and then taking the average of all reconstructed 
patches at position $(i,j)$. 
As seen in Figure~\ref{fig:patch_visual}, 
the patches along the border exhibit a systematically
higher RMSE. This is especially true at the corners which have the highest average RMSE. 
We speculate this is because patches at the edges and especially the corners are reconstructing with less information compared to patches towards the center. 
In essence, \enki\ is forced to extrapolate
instead of interpolate.
A distinct example of this can be see in Figure~\ref{fig:corner_example} where the gradient in the upper right corner is not as pronounced as the one found in the original image. In some cases, when a cutout has a small feature completely masked, the model will reconstruct the image without the feature entirely. 

While this issue may not be limited to edges and corners, Figure \ref{fig:corner_example} indicates that the edges and corners experience this issue more often than inner patches. 
A blunt solution to handling these patches is to exclude them altogether from the final cutout. While this throws out data, avoiding patches more prone to this issue 
may be critical for real-world applications.
For the remainder of this thesis,  we
ignore the border in all further calculations
and analysis.

Continuing our analysis of the patches, we also examined \enki's performance based on the 
complexity of a patch. 
Specifically, we gauged the patch complexity by
measuring the 
standard deviation in temperature (\stdT). 
We hypothesized that as patch complexity increases, 
we will also observe a rise in reconstruction RMSE. 
After binning patches by \stdT, we calculated
the median reconstruction error (RMSE) of each bin
(Figure~\ref{fig:patch_complexity}). Here, we plot the \stdT\ of a patch against the RMSE of a patch on a log-log scale. 

The black dashed line represents the one-to-one relationship between \stdT\ and RMSE. If the results for \enki\ track this line 
it would mean that the reconstructions are roughly equivalent
to random noise. The blue line is the RMSE of the $\tper=10$, $\pper=20$ patches plotted against the \stdT. The red dotted line represents a by-eye fit. 

We see that after a \stdT\ of 0.02K ($\approx -2$ on the plot) 
that the RMSE is smaller than the black line by nearly
one magnitude. We do see that there is a correlation between the RMSE and \stdT as they increase at a similar rate, but the lower RMSE means that the constructions being produced by \enki\ are excellent.

Before 0.02K, we observe that the plot flattens out at an RMSE of about 0.0012K (-3 on the plot). We suspect this is because as \stdT\ decreases, the special scale of features in the SST patterns also decreases. At the threshold of 0.02K, the spacial scale of the features become smaller than the size of the patch, and the reconstructed field within the patch becomes less reliant on information from outside the patch, resulting in patch reconstructions that are pretty much random. This indicates that as we surpass this threshold \enki's reconstructions start using outside information when reconstructing patches. 

\begin{figure}[ht]
\noindent\includegraphics[width=\textwidth]{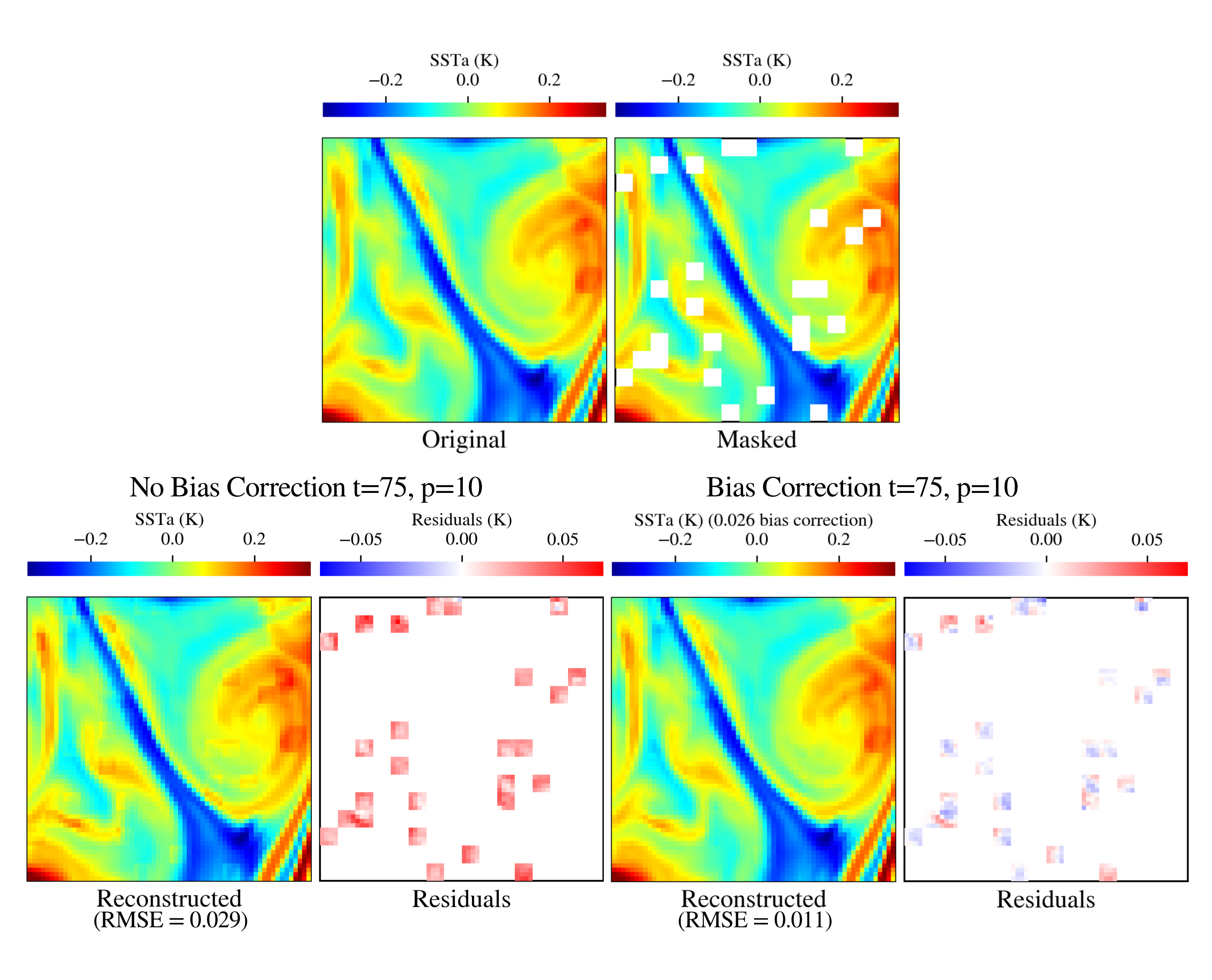}
\caption{An example of the offset that can be found in the $\tper=75$ model at $\pper=10$ and how adding a bias correction on the image can improve the final reconstruction. We see that in the example without bias that the RMSE is 0.0267 and that there is an offset in this image towards ~0.05K. In the 2nd image, we subtract the bias of 0.0267 from the reconstructed patches, and see an improvement in the RMSE down to 0.0162 and in the qualitative examination of the reconstruction.}
\label{fig:bias_example}
\end{figure}

\begin{figure}[ht]
\noindent\includegraphics[width=\textwidth]{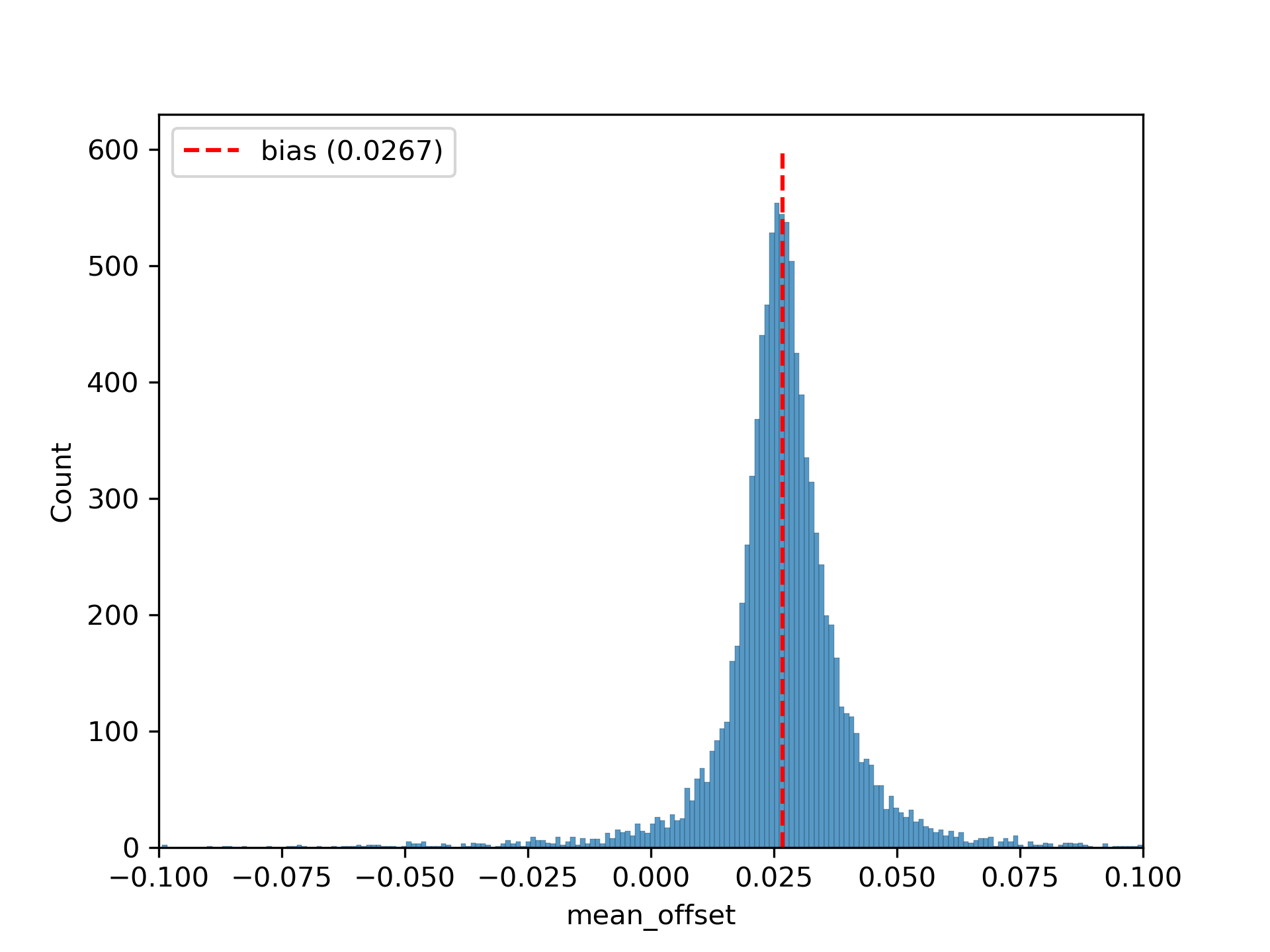}
\caption{
Histogram of the offsets in $\tper=75$ reconstructing $\pper=10$ images. The bias of the whole reconstruction set is 0.0267 which represented by the red line. 
}
\label{fig:offset_hist}
\end{figure}

\begin{figure}[ht]
\noindent\includegraphics[width=\textwidth]{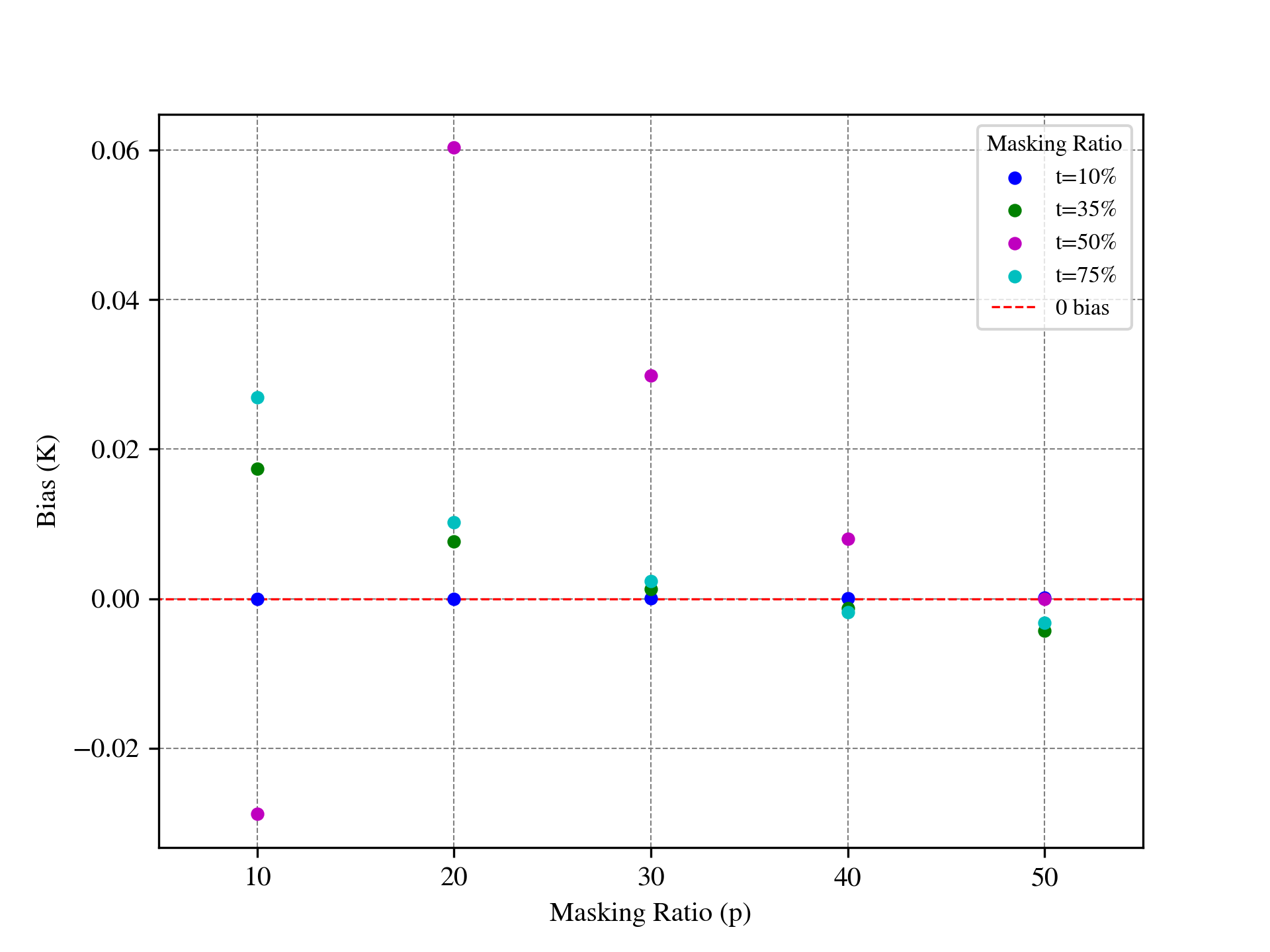}
% TODO
% 1. We can't refer to both t and p as "Masking Ratio".  Way too confusing.  How about Training Ratio for t and Patch Mask Ratio for p?   If you agree, make the changes throughout the thesis
% 2. Why put p on the x-axis?  That is, instead of t.  I think it would be more informative to do the opposite.
\caption{The bias that appears in $\tper=10,35,50,75$ at $\pper=10,20,30,40,50$. We mark zero bias as a dotted red line. In this plot, we see that in general, the $\tper=10$ model has a relatively low bias across all Patch Mask Ratio. The next lowest is the  $\tper=35$ which starts out with a higher bias until it reaches 3 $\pper=30$ masking. The  $\tper=75$ model follows a similar pattern to the  $\tper=35$ model. The  $\tper=50$ stands out as having a large bias at every \pper except  $\pper=50$.
}
\label{fig:model_bias}
\end{figure}

\subsection{Bias Analysis} 
\label{sec:bias}

When examining models trained with a higher masking ratio (\tper) that are reconstructing images of lower masking ratios 
(i.e. $\tper>\pper$), a systematic offset is 
observed (e.g.\ Figure~\ref{fig:offset_hist}).
This is most apparent when using the $\tper=75$ model to 
reconstruct images with $\pper=10$ and for the $\tper=50$ model 
aside from \pper=50. 
We calculate the offset in a single image as the median of the absolute value of the difference between original and reconstructed pixels. 
The mean offset of reconstructed set of images at 
a given \tper, \pper\ is referred to as the bias. 
For example, a single cutout reconstructed by the $\tper=75$ model at $\pper=10$ may have an offset of 0.0131, but the bias of the whole $\tper=75$,$\pper=10$ reconstruction set is 0.0269. When applying a correction to reconstructions, we will use the bias
because one cannot predict the correct offset for a given
cutout.

It is unclear why these offsets are present; we have yet to identify its origin or even
any strong correlation with image or patch properties.
But when the typical offsets are high enough, 
these lead to systematically large RMSE values.
We calculated the bias
for all models at all percentage of masking ratios, 
and plot them in Figure~\ref{fig:model_bias}. 
In addition, we provide the values
Table~\ref{tab:biases}. 

\begin{figure}[ht]
\noindent\includegraphics[width=\textwidth]{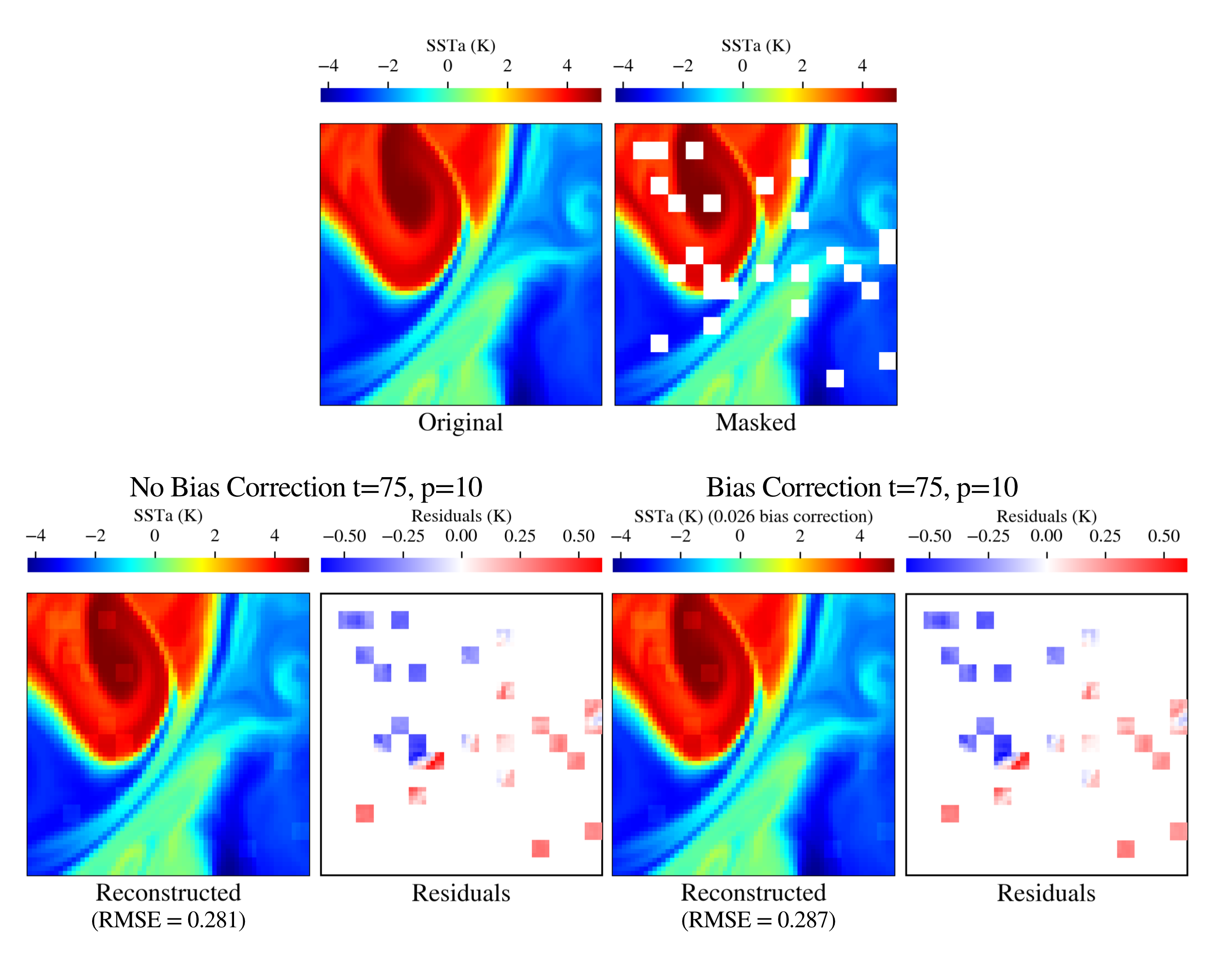}
\caption{In this example, despite subtracting the bias (0.026 for $\tper=75$,$\pper=10$) we see no improvement in the cutout. When we examine the RMSE, we see that adding the bias correction made the cutout worse. 
}
\label{fig:bias_example2}
\end{figure}

Figure \ref{fig:model_bias} shows that the most prominent bias 
occurs for models with high training mask ratios ($\tper > 10$)
applied to data with low masking ($\pper \le 20$).
When examining $\pper=10$, a bias is present in all of the models aside from the $\tper=10$ model. 
At $\pper=30$, we now see that the $\tper=35$ and $\tper=75$ model have a much lower bias, nearing the performance of the $\tper=10$ model. We do observe improvement in the $\tper=50$ model at $\pper=30$ as the bias increases by half from $\tper=20$ to $\tper=30$. 

As a whole, we note that the $\tper=10$ model's bias always remains close to zero while outside of the $\pper=50$. 

Qualitatively and quantitatively, 
adding the bias improves the reconstructions and lowers
the RMSE. Occasionally, the offset of an individual image is much larger than the bias, especially in images that have a larger \stdT. 
We show an example of this in Figure~\ref{fig:bias_example2} where despite adding a bias correction, the RMSE is worse than if we had not added the bias. When calculating the offset of this cutout, we find it to be 0.014. Despite this offset being half of what the bias is, the change is barely visible (if at all). 

%We find an answer to this when examining the residual which spans a range of -0.552K to 0.581K. Using our current method of calculating the offset, this means that the median value will fall closer to zero, meaning that correcting using this offset will not be effective. Additionally, this brings up an extra challenge when correcting the offset as we see that if there is an image with a large \stdT range that spans both positive and negative values, the offset will occur in both directions. Even if we were to correct with a value more tuned for this image, we will likely only correct either the positive values while making the negative values worse, or vise versa.

%Going forward, we will still correct images with the biases in Table \ref{tab:biases} because while not all reconstructions will be corrected, Figure \ref{fig:offset_hist} implies that by applying bias correction we will still be improving a majority of the cutouts. 

\begin{table}[]
\centering
\begin{tabular}{|l|c|c|c|c|c|}
\hline
Training Ratio &
  \multicolumn{1}{l|}{\textbf{$\pper=10$}} &
  \multicolumn{1}{l|}{\textbf{$\pper=20$}} &
  \multicolumn{1}{l|}{\textbf{$\pper=30$}} &
  \multicolumn{1}{l|}{\textbf{$\pper=40$}} &
  \multicolumn{1}{l|}{\textbf{$\pper=50$}} \\ \hline
\textbf{$\tper=10$} & 1.018e-0.6 & 1.214e-06 & 8.741e-0.6 & 8.027e-05 & 9.601e-05 \\ \hline
\textbf{$\tper=35$} & 0.0173     & 0.0077    & 0.0013     & -0.0013   & -0.0043   \\ \hline
\textbf{$\tper=50$} & -0.0288    & 0.0603    & 0.0298     & 0.00796   & -3.883e-6 \\ \hline
\textbf{$\tper=75$} & 0.0269     & 0.0102    & 0.00237    & -0.00183  & 0.00321   \\ \hline
\end{tabular}
\caption{The numerical values of all the biases. Notice how, for the $\tper=10,35,50,75$ models, the greater \pper\ is offset from the model's training ratio, the higher the bias. 
The only exception is the $\tper=10$ model, which has a low bias for all masking ratios.}
\label{tab:biases}
\end{table}

% %%%%%%%%%%%%%%%%%%%%%%%%%%%%%%%%%%%%%%%%%%%%%%%%%%%%%%%%%%%%%
\subsection{RMSE Analysis of Cutouts} 
\label{sec:rmse}

% This section should begin with:
%  3. Comment on the "absolute" performance of Enki, e.g. is it good enough for real-world applications? [oof]

\begin{figure}[ht]
\noindent\includegraphics[width=\textwidth]{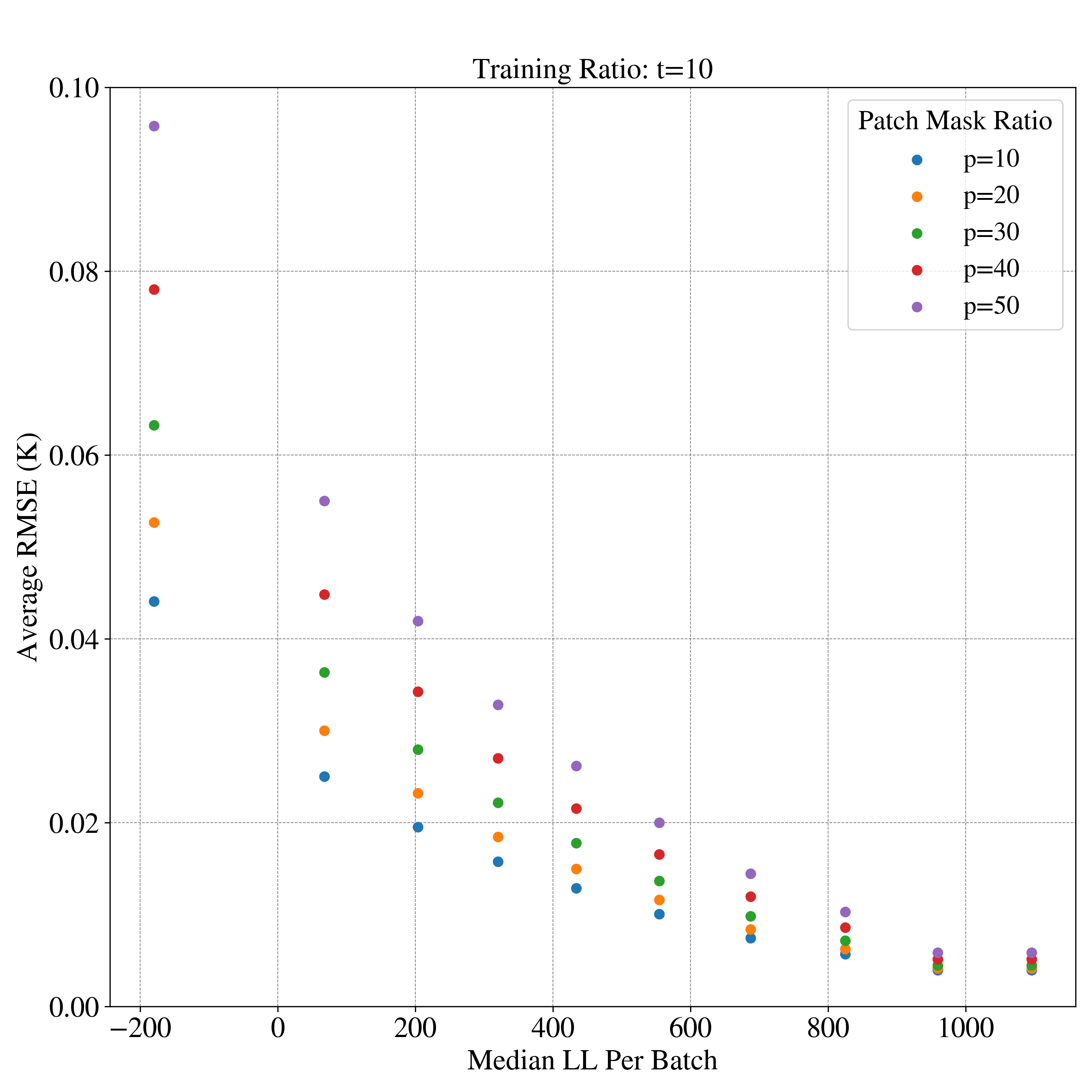}
\caption{
RMSE as a function of image complexity, here gauged by \LL\ as measured by
the \ulmo\ algorithm with higher values indicating less complexity.  
Irrespective of patch size, the average RMSE increase with image complexity
indicating the \enki\ model.  Aside from the lowest bin in \LL\ (most complex),
the RMSE of the reconstructed patches for $\pper \le 30$ have average RMSE~$<0.04$\,K
which is comparable to the noise in remote sensing sensors like VIIRS.
}
\label{fig:RMSE_vs_LL}
\end{figure}

\begin{figure}[ht]
\noindent\includegraphics[width=\textwidth]{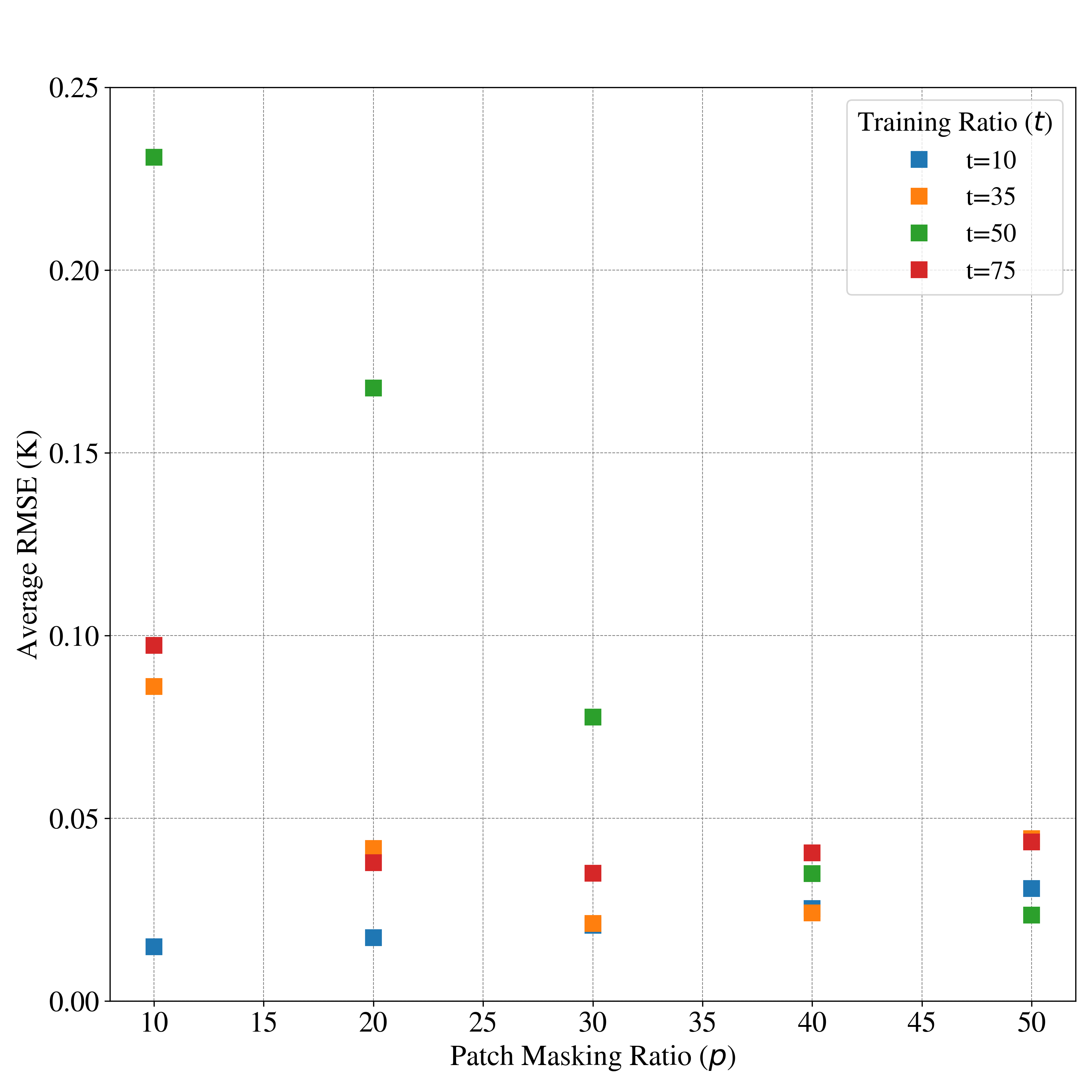}
\caption{
Average RMSE for all of the image reconstruction sets.
The \tper=10 model performs well at all masking ratios \pper.
}
\label{fig:RMSE_models}
\end{figure}

We hypothesize that recontruction performance will degrade
for images with higher complexity, e.g., sharp fronts.
%When examining the RMSE of cutouts, we chose to compare them to the original LL of the image to see whether complexity affected \enki's ability to reconstruct images. For this analysis, 
To test this, we remove the edge patches and apply a bias to all cutouts to ensure the best performance for all models. 
For image complexity, we adopt the \LL\ metric from \ulmo. 
%instead of \stdT\ because sorting by LL allows us to evaluate \enki's ability to reconstruct "common" vs "rare" SST patterns. 
%We performed this analysis across all \tper\ and \pper\ by taking the RMSE of all reconstructed cutouts and sorting them based on the LL of the original image. We then 
For each reconstruction set (\tper,\pper),
we `binned the RMSE into 10 groups of ascending LL and took the average of the RMSEs within those bins. We plot these results in Figures~\ref{fig:RMSE_vs_LL} and \ref{fig:RMSE_models}.

When comparing the RMSE to the log likelihood (\LL) we see there is a correlation between the two. As LL decreases, the RMSE increases. We also see that for the the masking ratio has an effect on how well a model performs. For $\tper=10$, as the masking ratio increases, the RMSE also increases. 

The initial observation from Figure \ref{fig:RMSE_models} is that the performance of the $\tper=50$ model was noticeably worse compared to the other models. This result matches the bad reconstructions we
have seen in Section~\ref{sec:decoder} and the higher biases in \ref{sec:bias}. This disparity can be attributed to the lower learning rate used during pre-training. We expect that increasing the number of training epochs or relaunching pre-training with a higher learning rate will make the $\tper=50$ model more competitive with the other models. Despite its poor showings at \pper, it is important to highlight that the $\tper=50$ model exhibited the best performance when reconstructing images at $\pper=50$. 

The $\tper=10$ model achieves the highest performance at $\pper=10$, followed by $\pper=20$ and then $\pper=30$. In the case of the $\tper=35$ model, the best performance is observed at $\pper=30$, followed by $\pper=40$ and $\pper=20$. The RMSE pattern of the $\tper=75$ model follows a slightly different trend. It exhibits its worst performance at $\pper=10$, shows improvement as it reaches $\pper=30$, and then experiences a gradual increase in RMSE to $\pper=50$. Although the exact cause for this behavior is uncertain, it is worth noting that the $\tper=75$ reconstructions across all the examined masking ratios have had issues. This suggests that the increasing masking ratio may contribute to the observed rise in RMSE after $\pper=30$.

Once again matching what we observed in Figure \ref{fig:model_comparison}, all models other than the $\tper=10$ model exhibit their worst performance at $\tper=10$. These are also the images that are observed to require the highest bias correction, and even with such correction, their performance remains more than twice as poor as that of the $\tper=10$ model.
Clearly, these models have not learned the finer-scale
features apparent in the data.

We acknowledge that the $\tper=10$ model consistently has some of the lowest RMSE. The only instances where we observe lower RMSE values than the $\tper=10$ model are with the $\tper=35$ model at $\pper=40$ and the $\tper=50$ model at $\pper=50$. The $\tper=35$ model does, however, compare competitively with the $\tper=10$ model at $\pper=30$ masking with the RMSE appearing very similar.

While this would indicate that the $\tper=10$ model would be a good general model to use for reconstructions, it is important to remember that when examining the $\tper=10$ models qualitatively, we can see that the low RMSE values are likely because it is able to estimate the values of the image well even at higher masking ratios. 
%This however does not mean that the reconstructions are perfect. This is most apparent when we look at $\tper=10$ cutouts at these higher masking ratios. In Figure \ref{fig:model_comparison} we saw that the $\tper=10$ model started seeing some signs of degradation in reconstruction at as early as $\pper=30,50$. Comparatively, we do not see these same issues with the patches when looking at the at the $\tper=35$,$\pper=35$ and $\tper=50$,$\pper=50$ reconstructions.

%We looked at more $\tper=10$, $\pper=40$ reconstructions because $\pper=40$ is the first masking ratio that another model outperforms the $\tper=10$ model, and we are able to find many examples of bad reconstructions, and in Figure \ref{fig:RMSE_models} we show some of these failed reconstructions. 
%In the first and third cutouts, $\tper=10$ completely fails to correctly guess even the values or the dynamics of the cutout. 
%In the second example, while the reconstruction values are much closer to the original image, they are still not correct, and the patches appear to have poorly reconstructed the dynamics of the original image. This observation supports our theory that there is a threshold at which a model performs optimally.

Overall \enki\ shows great promise in being able reconstruct SST images. While some of the finer details can be lost especially in higher masking ratios, we have seen countless examples in which \enki\ is still able to capture the general structure of the cutout. In addition, for all masking ratios we examined, there was always a model that had an RMSE lower than 0.05. We believe that \enki\ shows great potential for reconstructing clouds masks in real SST images.

\begin{figure}[ht]
\noindent\includegraphics[width=\textwidth]{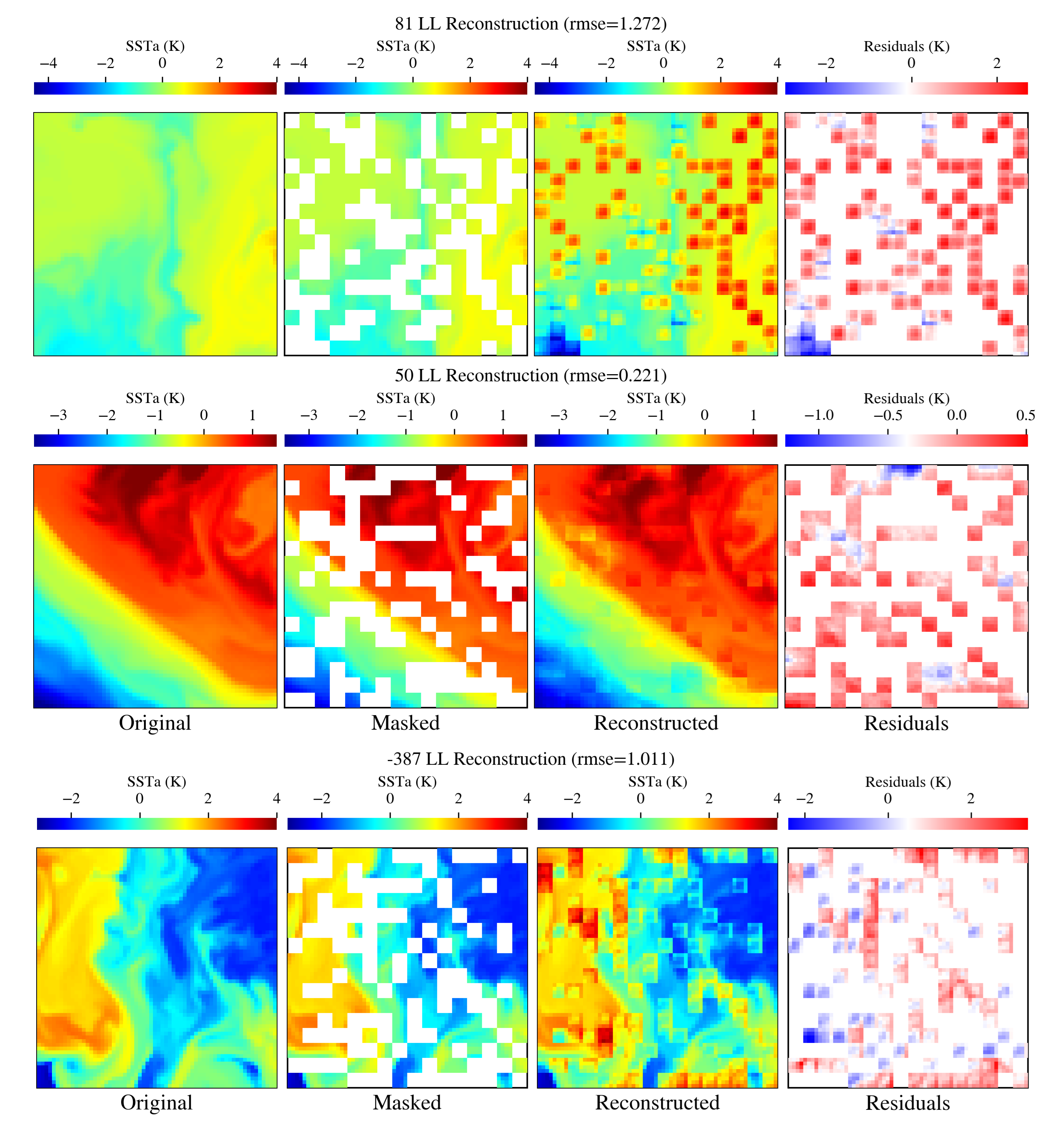}
\caption{
Different examples of how the $\tper=10$ model incorrectly reconstructs $\pper=40$ cutouts. In the first, $\tper=10$ completely fails to guess the values of the original image and fills in the patch with extreme values. In the second example, while performing better than the first example, the patches are still visible. In the third example, we see that $\tper=10$ has failed to both capture the values and the structure of the patterns with patches have a blocky characteristic to them. 
}
\label{fig:t10_p40_fails}
\end{figure}

\section{Concluding Remarks}

Our analysis reveals that \enki\ shows great promise as a method of reconstructing cloud masked SST images. The qualitative examinations produce sensible and usable results, and by exploring various training ratios, we aim to identify the ratio that produces the most favorable and effective outcomes.

We find \enki\ is most accurate when handling reconstructions at lower masking ratios with higher masking ratios running the risk of masking out too much information which may lead \enki\ to smoothing over or missing masked features. We also note, that patches at the edge of cutouts are more prone to error, leading us to conclude that when reconstructing, it's best to remove the edge patches to reduce the chance of errors.

We do, however, acknowledge that there our some limitations in our current analysis. The "clouds" we introduce into images are randomized, so they are not a perfect representation of how real cloud masks will look. Going forward, it would be good to test how \enki\ performs reconstructing LLC cutouts with real cloud masks. Throughout this thesis, we also mainly focus on LLC data which is free of a lot of issues that plague real data. Real remote sensing data is noisy because of atmospheric noise and noise introduced by the instruments, and it often runs the risk of unmasked clouds. The next step would be to run \enki\ on real world data and see how it compares to older methods of filling masked pixels like inpainting. 

We also acknowledge that while \enki\ appears to be able to generalize well and capture the main shapes of a dynamic, there is a lot of information that it is unable to recreate accurately at a $\pper=75$ masking ratio. This becomes apparent when comparing the RMSE of reconstructions at different masking percentages where we see a steady increase in the RMSE as the masking ratio increases. This means that while the models like the $\tper=75$ model can reconstruct up to $\pper=75$ masking, we should be cautious because reconstructions at this level of \pper\ masking lose much of the details due to aggressive masking.

If we were to only use a single model, it would likely be the $\tper=10$ model. Reconstructions would likely only go up to $\pper=20$, which if applied to VIIRS data would increase the available cutouts by a magnitude (Figure \ref{fig:cloud_coverage}). However, since we know that different models perform best within a range of \pper, an alternative approach is to incorporate multiple $\tper$ models within \enki\ and select the appropriate model based on the cloud coverage of the cutout being reconstructed. To determine the exact range where each model performs optimally, further in-depth analysis is required.

Future work will encompass a range of tasks and objectives.
\begin{itemize}
\item Our subsequent undertaking involves conducting analysis on VIIRS cutouts. While the reconstructions of LLC cutouts displayed promising results, it is crucial to determine if these outcomes can be effectively applied to real data. This is because, despite the well-predicted SST patterns of VIIRS determined by the LLC4320 model, it is important to acknowledge that the LLC4320 data itself is generated and may have variations when compared to real-world observations.
\item Moving forward, we also aim to conduct reconstructions utilizing real cloud masks obtained from actual data. The current random masking, especially at lower masking ratios, may not accurately represent the true cloud masking scenario. Therefore, applying real cloud masks in our reconstructions has the potential to yield different and more realistic results, providing valuable insights into the performance of the reconstruction model.
\item Throughout this study, our training dataset and validation set consisted of noise-free cutouts. While the primary objective of \enki\ is not to reconstruct noise, we want to investigate whether the inclusion of noise has any impact on the quality of reconstructions. By introducing noise into the input data, we can assess the robustness and performance of \enki\ under more realistic conditions.
\item Additionally, we consider conducting a follow-up study in which we set the seed of the mask for reconstructions. This controlled experiment will enable us to analyze the specific impact of each model on the reconstructed output, without confounding factors introduced by random variations.
\item We may explore the use of different types of loss functions during training or alternative equations to evaluate the performance of reconstructions. It is worth considering that many images exhibit significant variations in stdT, which means that the perception of reconstruction errors may differ based on the magnitude of stdT. To address this, we could potentially evaluate a model's performance by calculating the RMSE normalized by the stdT of the masked pixels in the original image. This approach would provide a more contextually relevant assessment of the reconstruction quality, accounting for the inherent variability in sea surface temperature across different images.

\end{itemize}

\section{Acknowledgements}

AA acknowledges J. Xavier Prochaska and P. C.
Cornillon’s invaluable guidance on this project, as well as
David Reiman’s and Edwin Goh's insightful discussions.
The authors acknowledge use of the Nautilus cloud computing system which is supported by the following US National Science Foundation (NSF) awards: CNS-1456638, CNS1730158, CNS-2100237, CNS-2120019, ACI-1540112, ACI1541349, OAC-1826967, OAC-2112167.

%%%%%%%%%%%%%%%%%%%%%%%%%%%%%%%%%%%%%%%%%%%%%%%%%%%%%%%%%%%%%%%%
%
%  ACKNOWLEDGMENTS
%
% The acknowledgments must list:
%
% >>>>	A statement that indicates to the reader where the data
% 	supporting the conclusions can be obtained (for example, in the
% 	references, tables, supporting information, and other databases).
%
% 	All funding sources related to this work from all authors
%
% 	Any real or perceived financial conflicts of interests for any
%	author
%
% 	Other affiliations for any author that may be perceived as
% 	having a conflict of interest with respect to the results of this
% 	paper.
%
%
% It is also the appropriate place to thank colleagues and other contributors.
% AGU does not normally allow dedications.

%% ------------------------------------------------------------------------ %%
%% References and Citations

%%%%%%%%%%%%%%%%%%%%%%%%%%%%%%%%%%%%%%%%%%%%%%%
%
% \bibliography{<name of your .bib file>} don't specify the file extension
%
% don't specify bibliographystyle
%%%%%%%%%%%%%%%%%%%%%%%%%%%%%%%%%%%%%%%%%%%%%%%

%\bibliography{copernicus}
\bibliographystyle{copernicus}
\bibliography{references.bib}

\begin{thebibliography}{15}
\providecommand{\natexlab}[1]{#1}
\providecommand{\url}[1]{{\tt #1}}
\providecommand{\urlprefix}{URL }
\expandafter\ifx\csname urlstyle\endcsname\relax
  \providecommand{\doi}[1]{https://doi.org/\discretionary{}{}{}#1}\else
  \providecommand{\doi}{https://doi.org/\discretionary{}{}{}\begingroup
  \urlstyle{rm}\Url}\fi

\bibitem[{Baldi(2012)}]{baldi2012autoencoders}
Baldi, P.: Autoencoders, unsupervised learning, and deep architectures, in:
  Proceedings of ICML workshop on unsupervised and transfer learning, pp.
  37--49, JMLR Workshop and Conference Proceedings, 2012.

\bibitem[{{Bertalmio} et~al.(2001){Bertalmio}, {Bertozzi}, and
  {Sapiro}}]{bertalmio+2001}
{Bertalmio}, M., {Bertozzi}, A.~L., and {Sapiro}, G.: Navier-stokes, fluid
  dynamics, and image and video inpainting, in: Proceedings of the 2001 IEEE
  Computer Society Conference on Computer Vision and Pattern Recognition. CVPR
  2001, vol.~1, pp. I--I, \doi{10.1109/CVPR.2001.990497}, 2001.

\bibitem[{Chelton et~al.(2011)Chelton, Schlax, and Samelson}]{eddies}
Chelton, D.~B., Schlax, M.~G., and Samelson, R.~M.: Global observations of
  nonlinear mesoscale eddies, Progress in Oceanography, 91, 167--216,
  \doi{https://doi.org/10.1016/j.pocean.2011.01.002}, 2011.

\bibitem[{Dosovitskiy et~al.(2020)Dosovitskiy, Beyer, Kolesnikov, Weissenborn,
  Zhai, Unterthiner, Dehghani, Minderer, Heigold, Gelly, Uszkoreit, and
  Houlsby}]{vit}
Dosovitskiy, A., Beyer, L., Kolesnikov, A., Weissenborn, D., Zhai, X.,
  Unterthiner, T., Dehghani, M., Minderer, M., Heigold, G., Gelly, S.,
  Uszkoreit, J., and Houlsby, N.: An Image is Worth 16x16 Words: Transformers
  for Image Recognition at Scale, CoRR, abs/2010.11929,
  \urlprefix\url{https://arxiv.org/abs/2010.11929}, 2020.

\bibitem[{{Gallmeier} et~al.(2023){Gallmeier}, {Prochaska}, {Cornillon},
  {Menemenlis}, and {Kelm}}]{ulmo_on_llc}
{Gallmeier}, K., {Prochaska}, J.~X., {Cornillon}, P.~C., {Menemenlis}, D., and
  {Kelm}, M.: {An evaluation of the LLC4320 global ocean simulation based on
  the submesoscale structure of modeled sea surface temperature fields}, arXiv
  e-prints, arXiv:2303.13949, \doi{10.48550/arXiv.2303.13949}, 2023.

\bibitem[{He et~al.(2021)He, Chen, Xie, Li, Doll{\'{a}}r, and
  Girshick}]{vitmae}
He, K., Chen, X., Xie, S., Li, Y., Doll{\'{a}}r, P., and Girshick, R.~B.:
  Masked Autoencoders Are Scalable Vision Learners, CoRR, abs/2111.06377,
  \urlprefix\url{https://arxiv.org/abs/2111.06377}, 2021.

\bibitem[{{Isern-Fontanet} et~al.(2017){Isern-Fontanet}, {Ballabrera-Poy},
  {Turiel}, and {Garc{\'\i}a-Ladona}}]{sst_examples}
{Isern-Fontanet}, J., {Ballabrera-Poy}, J., {Turiel}, A., and
  {Garc{\'\i}a-Ladona}, E.: {Remote sensing of ocean surface currents: a review
  of what is being observed and what is being assimilated}, Nonlinear Processes
  in Geophysics, 24, 613--643, \doi{10.5194/npg-24-613-2017}, 2017.

\bibitem[{Jonasson and Ignatov(2019)}]{viirs}
Jonasson, O. and Ignatov, A.: {Status of second VIIRS reanalysis (RAN2)}, in:
  Ocean Sensing and Monitoring XI, edited by Hou, W.~W. and Arnone, R.~A., vol.
  11014, p. 110140O, International Society for Optics and Photonics, SPIE,
  \doi{10.1117/12.2518908}, 2019.

\bibitem[{Levy et~al.(2018)Levy, Franks, and Smith}]{currents}
Levy, M., Franks, P., and Smith, K.: The role of submesoscale currents in
  structuring marine ecosystems, Nature Communications, 9,
  \doi{10.1038/s41467-018-07059-3}, 2018.

\bibitem[{McWilliams(2017)}]{fronts}
McWilliams, J.~C.: Submesoscale surface fronts and filaments: secondary
  circulation, buoyancy flux, and frontogenesis, Journal of Fluid Mechanics,
  823, 391–432, \doi{10.1017/jfm.2017.294}, 2017.

\bibitem[{Penven et~al.(2006)Penven, Debreu, Marchesiello, and
  McWilliams}]{upwelling}
Penven, P., Debreu, L., Marchesiello, P., and McWilliams, J.~C.: Evaluation and
  application of the ROMS 1-way embedding procedure to the central california
  upwelling system, Ocean Modelling, 12, 157--187,
  \doi{https://doi.org/10.1016/j.ocemod.2005.05.002}, 2006.

\bibitem[{{Prochaska} et~al.(2021){Prochaska}, {Cornillon}, and
  {Reiman}}]{ulmo}
{Prochaska}, J.~X., {Cornillon}, P.~C., and {Reiman}, D.~M.: {Deep Learning of
  Sea Surface Temperature Patterns to Identify Ocean Extremes}, Remote Sensing,
  13, 744, \doi{10.3390/rs13040744}, 2021.

\bibitem[{{Prochaska} et~al.(2023){Prochaska}, {Beaulieu}, and
  {Giamalaki}}]{pbg2023}
{Prochaska}, J.~X., {Beaulieu}, C., and {Giamalaki}, K.: {The rapid rise of
  severe marine heat wave systems}, Environmental Research: Climate, 2, 021002,
  \doi{10.1088/2752-5295/accd0e}, 2023.

\bibitem[{Prochaska et~al.(2023)Prochaska, Guo, Cornillon, and
  Buckingham}]{ocean_patterns}
Prochaska, J.~X., Guo, E., Cornillon, P.~C., and Buckingham, C.~E.: The
  Fundamental Patterns of Sea Surface Temperature, 2023.

\bibitem[{Wu et~al.(2017)Wu, Cornillon, Boussidi, and Guan}]{cloud_noise}
Wu, F., Cornillon, P., Boussidi, B., and Guan, L.: Determining the
  Pixel-to-Pixel Uncertainty in Satellite-Derived SST Fields, Remote Sensing,
  9, \urlprefix\url{https://www.mdpi.com/2072-4292/9/9/877}, 2017.

\end{thebibliography}
%\printbibliography

%Reference citation instructions and examples:
% 
% Please use ONLY \cite and \citeA for reference citations.
% \cite for parenthetical references
% ...as shown in recent studies (Simpson et al., 2019)
% \citeA for in-text citations
% ...Simpson et al. (2019) have shown...
%
%
%...as shown by \citeA{jskilby}.
%...as shown by \citeA{lewin76}, \citeA{carson86}, \citeA{bartoldy02}, and \citeA{rinaldi03}.
%...has been shown \cite{jskilbye}.
%...has been shown \cite{lewin76,carson86,bartoldy02,rinaldi03}.
%... \cite <i.e.>[]{lewin76,carson86,bartoldy02,rinaldi03}.
%...has been shown by \cite <e.g.,>[and others]{lewin76}.
%
% apacite uses < > for prenotes and [ ] for postnotes
% DO NOT use other cite commands (e.g., \citet, \citep, \citeyear, \nocite, \citealp, etc.).
%

\end{document}